\documentclass[manuscript]{acmart}
\usepackage{graphicx}
\usepackage{subcaption}
\usepackage{algpseudocode}
\usepackage{algorithm}
\usepackage{tcolorbox}
\usepackage{comment}
\usepackage{tcolorbox}
\AtBeginDocument{%
  }

\setcopyright{acmlicensed}
\copyrightyear{2025}
\acmYear{2025}
\acmDOI{XXXXXXX.XXXXXXX}

\acmJournal{TOSEM}
\acmVolume{37}
\acmNumber{4}
\acmArticle{111}
\acmMonth{8}

\begin{document}

\title{Representation Improvement in Latent Space for Search-Based Testing of Autonomous Robotic Systems}

\author{Dmytro Humeniuk}
\email{dmytro.humeniuk@polymtl.ca}
\orcid{0000-0002-2983-8312}
\affiliation{%
  \institution{Polytechnique Montréal}
  \city{Montreal}
  \state{QC}
  \country{Canada}
}

\author{Foutse Khomh}
\affiliation{%
  \institution{Polytechnique Montréal}
  \city{Montreal}
  \state{QC}
  \country{Canada}
}

\newcommand{\Foutse}[1]{\textcolor{red}{{\it [Foutse says: #1]}}}
\newcommand{\Dima}[1]{\textcolor{blue}{{ [#1]  }}}
\renewcommand{\shortauthors}{Humeniuk D. and Khomh F.}
\acmArticleType{Review}
\acmCodeLink{https://github.com/swat-lab-optimization/RILaST}
\acmDataLink{https://zenodo.org/records/15087028}


\begin{abstract}
Testing autonomous robotic systems, such as self-driving cars and unmanned aerial vehicles, is challenging due to their interaction with highly unpredictable environments. A common practice is to first conduct simulation-based testing, which, despite reducing real-world risks, remains time-consuming and resource-intensive due to the vast space of possible test scenarios. A number of search-based approaches were proposed to generate test scenarios more efficiently. A key aspect of any search-based test generation approach is the choice of representation used during the search process. However, existing methods for improving test scenario representation remain limited. We propose RILaST (Representation Improvement in Latent Space for Search-Based Testing) approach, which enhances test representation by mapping it to the latent space of a variational autoencoder. We evaluate RILaST on two use cases, including autonomous drone and autonomous lane-keeping assist system. The obtained results show that RILaST allows finding between 3 to 4.6 times more failures than baseline approaches, achieving a high level of test diversity.

\end{abstract}


\maketitle

\section{Introduction}
The architectures of modern autonomous robotic systems (ARS), such as self-driving cars, are highly complex, integrating hardware, software, and machine learning components that interact in dynamic and often unpredictable ways. Consequently, ensuring the robustness of ARS software is essential to enable these systems to withstand the numerous challenges present in unstructured real-world environments.

Simulation-based testing is a widely adopted method for evaluating ARS performance, where system models are assessed under various scenarios within a virtual environment. However, these simulations are computationally expensive, since each run can take from tens of seconds to several minutes. Given the limited testing time budget, selecting the most promising simulations is critical. Random test selection typically generates many irrelevant or non-challenging scenarios, thereby reducing efficiency. To overcome this limitation, several search-based approaches have been proposed for more efficient test generation~\cite{sorokin2024guiding, zohdinasab2024focused, khatiri2023simulation, haq2022efficient}. The effectiveness of these approaches is heavily influenced by the choice of representation, which should facilitate the discovery of challenging test scenarios while promoting a diverse set of tests~\cite{gaier2020discovering}.

Recent techniques in autonomous systems testing have specifically targeted the need for diverse scenario representation. For example, the Doppelganger tool~\cite{huai2023doppelganger} for testing autonomous driving systems (ADS) specifies test cases by detailing the positions and trajectories of autonomous vehicles (AVs) and pedestrians, as well as dynamic traffic signal configurations. In Doppelganger, each AV is represented by a set of 2D points outlining its expected path along with a movement start timestep, while each pedestrian is characterized by a set of 2D points, a movement start time, and a speed. Traffic control configurations are defined by five parameters that specify the signal color at each moment of the simulation. Although this expressive representation enables the specification of a vast number of test cases, it also leads to increased complexity in the scenario search space. Given the non-negligible time required for each simulation, efficiently exploring this complex space—even with search-based techniques—remains a significant challenge.

Concurrently, recent work in the evolutionary search community~\cite{volz2018evolving, bontrager2018deep, bentley2022evolving} has demonstrated the benefits of leveraging the latent space of generative models, such as generative adversarial networks (GANs)~\cite{goodfellow2014generative} and variational autoencoders (VAEs)~\cite{kingma2013auto}, for optimization tasks. The latent space offers a smoother, more continuous search landscape and encodes a bias toward high-performing solutions derived from the data. Intuitively, these approaches transform a difficult-to-search genotype space into a more structured and navigable latent space.

To the best of our knowledge, no approach in the search-based testing literature has yet explored mapping the genotype search space to a latent space to improve representation. To address this gap, we propose Representation Improvement in Latent Space for Search Based Testing of Autonomous Robotic Systems (RILaST), a framework that leverages evolutionary search in conjunction with variational autoencoders to optimize the search over a latent space populated with challenging and diverse test scenarios. 
\begin{figure}[h!]
\begin{subfigure}{0.45\textwidth}
  \centering
  \includegraphics[scale=0.4]{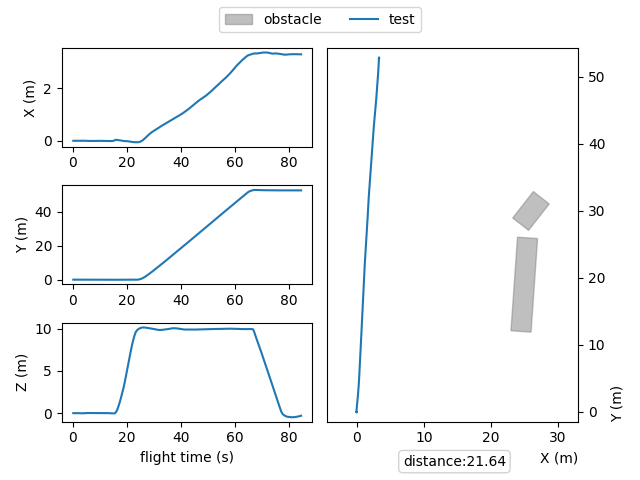}  
  \caption{Test with obstacles far from the UAV path}
  \label{fig:uav_scenario1}
\end{subfigure}
\begin{subfigure}{0.4\textwidth}
  \centering
  \includegraphics[scale=0.4]{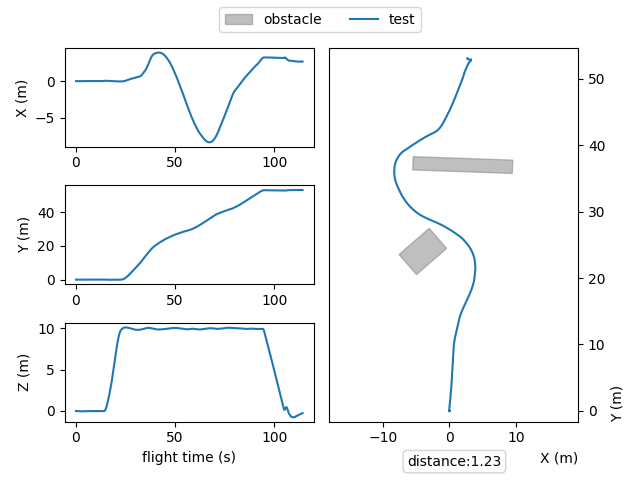}  
  \caption{Test with obstacles close to the UAV path}
  \label{fig:uav_scenario2}
\end{subfigure}
\caption{Examples of non-failing and failing test scenarios for the UAV system}
\label{fig:test_scenario_uav}
\end{figure}

To illustrate the potential benefits of learning a latent space for test generation, consider an unmanned aerial vehicle (UAV) equipped with an obstacle avoidance system. The UAV is required to complete a mission by flying straight for 50 meters while maintaining a distance of at least 1.5 meters from any obstacle. As shown in Figure~\ref{fig:uav_scenario1}, a test scenario with obstacles located far from the UAV's flight path is less likely to induce a failure, making it less valuable for testing. Conversely, applying a simple heuristic—such as positioning obstacles closer to the UAV's path—can significantly increase the difficulty of the test scenario, as depicted in Figure~\ref{fig:uav_scenario2}, where the UAV fails to meet the safety requirement. However, explicitly incorporating this heuristic as an additional constraint may lead to a more discontinuous search space, complicating the search process~\cite{rothlauf2002representations}. This motivates our approach of optimizing test generation within the latent space, where an improved representation can be learned from data containing diverse, failure-revealing tests. Our experiments on two use cases demonstrate that this improved representation enables the search algorithm to navigate the space more efficiently, uncovering 3–4.5 times more diverse failure-revealing scenarios compared to searches in the original space.

In this paper, we make the following contributions: 
\begin{enumerate}
    \item RILaST, a search-based framework that enhances representation for search-based testing by learning it in the latent space of a variational autoencoder;
    \item An extensive evaluation of RILaST and comparison to baselines, including state-of-the art tools and alternative implementations of the search algorithm used in RILaST;
    \item A publicly available replication package with the implementation of RILaST~\cite{humeniuk_2025_15087028}.
\end{enumerate}
The remainder of this paper is organized as follows. Section~\ref{sec:background} provides background on variational autoencoders and genetic algorithms (GA). In Section~\ref{sec:problem}, we formalize the problem of scenario generation and describe the specific test generation challenges addressed in this work. Section~\ref{sec:RILaST} details the RILaST approach and our methodology. In Section~\ref{sec:evaluation}, we present our research questions, study subjects, evaluation results, and potential threats to validity. Section~\ref{sec:discussion} discusses the benefits and challenges associated with learning the latent representation for test generation, as well as avenues for future work. Section~\ref{sec:literature} reviews related work in representation improvement for evolutionary search and autonomous systems testing. Finally, Section~\ref{sec:conclusions} concludes the paper.

\section{Background}\label{sec:background}
In this section, we provide background information on generative models, particularly variational autoencoders, which learn a stochastic mapping between samples in the latent space and the original data space~\cite{kingma2013auto}. We then describe the process of evolutionary search, which we use to guide the test generation in the latent space.
\subsection{Variational autoencoders}
Generative models, such as autoencoders, generative adversarial networks, and variational autoencoders, rely on latent variables sampled from a latent space to learn and approximate the original data distribution. By sampling from this distribution, they can generate new data similar to the original dataset. In other words, they learn a probabilistic representation of the input data. 
At the same time, GANs and autoencoders have a different underlying training methodology. In GANs, two modules need to be trained, namely the generator and the discriminator, which have opposite goals. The generator should learn to produce data that the discriminator cannot differentiate from the original data distribution. Because of this complex training dynamics, GANs have well-known challenges in training, such as mode collapse and instability~\cite{adler2018banach}. Autoencoders, by contrast, optimize a well-defined objective function by minimizing the reconstruction loss between input and output data, which enhances their stability and makes them easier to train~\cite{kingma2019introduction}. 

Vanilla autoencoders~\cite{Goodfellow-et-al-2016} encode the input into a fixed latent vector $z$, which is more suitable for data compression than for generation. During training, they are designed to minimize the reconstruction loss (e.g., mean squared error or binary cross-entropy), which is effective for memorizing the inputs rather than learning a useful and generalizable representation of the data. This limitation makes vanilla autoencoders less suitable for our task of representation improvement.

At the same time, VAEs learn a probabilistic distribution over the latent space, which enables better generalization and meaningful data interpolation. Unlike vanilla autoencoders, which can overfit to the training data, VAEs introduce a regularization term that encourages smooth and continuous latent representations. During training, they minimize a combined loss: a reconstruction loss to ensure the decoded outputs resemble the inputs and a Kullback-Leibler (KL) divergence term to regularize the latent space, preventing overfitting and promoting meaningful structure in the learned representation. This makes VAEs better suited for representation improvement tasks. Therefore, we chose VAEs for further experiments.

\begin{figure}[h!]
\begin{subfigure}{0.45\textwidth}
  \centering
  \includegraphics[scale=0.4]{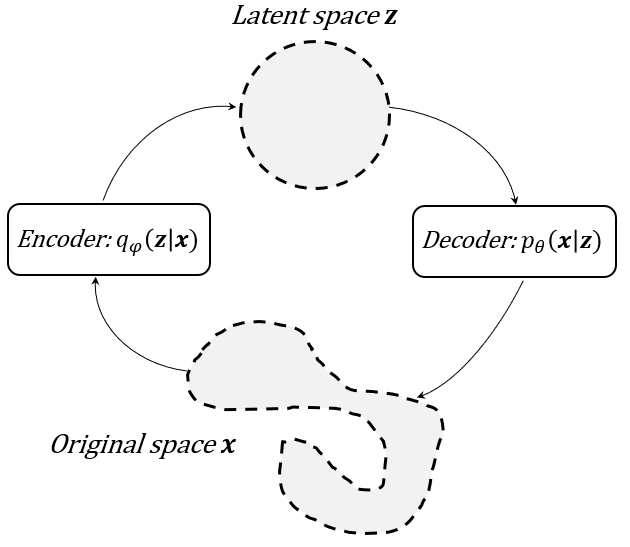}  
  \caption{VAE concept diagram}
  \label{fig:vae1}
\end{subfigure}
\begin{subfigure}{0.45\textwidth}
  \centering
  \includegraphics[scale=0.52]{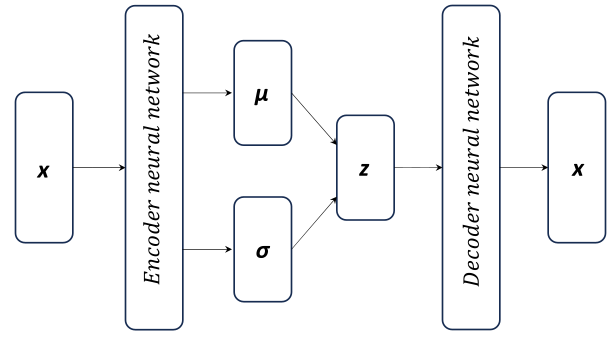}  
  \caption{VAE implementation diagram}
  \label{fig:vae2}
\end{subfigure}
\caption{VAE diagrams}
\label{fig:vae}
\end{figure}

As shown in Figure~\ref{fig:vae1}, the architecture of a VAE consists of two main components: encoder and decoder. 
The encoder learns a distribution $q_{\phi}(z|x)$ that approximates the true posterior $p(z|x)$, mapping data points into a structured latent space. This latent space captures the essential variations in the data, allowing for smooth interpolations and meaningful representations. The decoder, defined by $p_{\theta}(x|z)$, learns to generate realistic outputs from latent samples.
Training a VAE involves optimizing the Evidence Lower Bound (ELBO), which balances two objectives. The first term ensures that the decoder can accurately reconstruct the input, encouraging meaningful representations in the latent space. The second term regularizes the latent distribution by enforcing it to be close to a prior distribution $p(z)$, typically a standard normal distribution $\mathcal{N}(0, I)$. This prevents overfitting and ensures smoothness in the latent space. The ELBO loss function is given by:  

\begin{align}
\mathcal{L}(\theta, \phi) = \mathbb{E}_{q_{\phi}(z|x)} \left[ \log p_{\theta}(x|z) \right] - D_{\text{KL}}(q_{\phi}(z|x) \parallel p(z))
\end{align}
The first term represents the expected log-likelihood of reconstructing $x$ from the latent variable $z$, where $z$ is sampled from the learned latent distribution $q_{\phi}(z \mid x)$. 
 This term acts as a reconstruction loss, encouraging the model to generate an output that closely matches the original input $x_i$ from its latent representation $z_i$.
In practice, this term is often computed using the Mean Squared Error (MSE) for continuous data or Binary Cross-Entropy (BCE) for binary data. The second term is the KL divergence, which encourages the learned latent distribution $q_{\phi}(z|x)$ to be close to the prior $p(z)$.

Figure~\ref{fig:vae2} illustrates the architecture of a VAE, including its key components. The process begins with an input \( x \), which is passed through an encoder, which is typically implemented as a neural network (NN). The encoder maps \( x \) to a latent distribution, parameterized by a mean \( \mu \) and a standard deviation \( \sigma \), rather than directly to a fixed latent vector. The latent variable \( z \) is then sampled from this distribution. The sampled latent representation \( z \) is then passed through the decoder,  which reconstructs an approximation of the original input \( x \).

\subsection{Evolutionary search}
\begin{figure}[h!]
\includegraphics[scale=0.46]{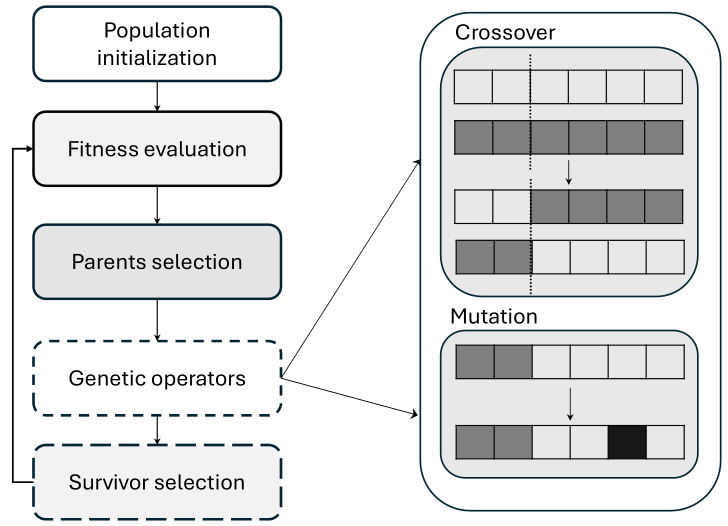}
\centering
\caption{Evolutionary search pipeline}
\label{fig:ga_pipeline}
\end{figure}

Genetic algorithms are a class of evolutionary algorithms that use a set of principles from genetics, such as selection, crossover, and mutation, to evolve a population of candidate solutions~\cite{back1997handbook}.
They have been previously actively used in software engineering community for test generation tasks~\cite{harman2008search}.
A typical genetic algorithm pipeline is shown in Figure~\ref{fig:ga_pipeline}.
The basic idea is to start with a set of individuals (candidate solutions) representing the initial population, usually generated randomly from the allowable range of values. Each individual is encoded in a dedicated form, such as a vector of real numbers, and is called a chromosome. A chromosome is composed of genes. Each individual is evaluated and assigned a fitness value. Some of the individuals are selected for mating. The search is continued until a stopping criterion is satisfied or the number of iterations exceeds a specified limit. 
Three genetic operators are used to evolve the solutions: selection of survivors and parents, crossover, and mutation.
The operators in evolutionary algorithms serve different functions. Mutation and crossover operators promote diversity in the population, allowing for the exploration of the solution space. On the other hand, survivor and parent selection operators promote the quality of the solutions, enabling the exploitation of the search space. It is essential to select the appropriate combination of operators that offers a good balance between exploration and exploitation to achieve optimal results~\cite{eiben2015introduction}.

Selection of parents is an operator that gives solutions with higher fitness a higher probability of contributing to one or more children in the succeeding generation. The intuition is to give better individuals more opportunities to produce offspring.
One of the commonly used selection operators is  
tournament selection~\cite{goldberg1991comparative}. The crossover operator is used to exchange characteristics of candidate solutions among themselves. The mutation operator has been introduced to prevent convergence to local optima; it randomly modifies an individual’s genome (e.g., by flipping some of its bits, if the genome is represented by a bitstring)~\cite{antoniol2005search}. In contrast to parent selection, which is typically
stochastic, survivor selection is often deterministic.
For genetic algorithms ($\mu$ + $\lambda$), the survivor selection strategy is commonly used. In this strategy, the set of offspring and parents are merged and ranked according to (estimated) fitness, then the top $\mu$ individuals are kept to form the next generation
~\cite{eiben2015introduction}.

\section{Problem formulation}\label{sec:problem}

In this section, we formulate the problem of generating test scenarios for autonomous robotic systems as an optimization problem. We then provide a detailed description of this problem in the context of two case studies: autonomous lane keeping and autonomous navigation.

\subsection{Test scenario generation problem}

We begin by formally defining the problem of test scenario generation as an optimization problem. This formulation includes constraints to ensure the generation of valid test scenarios and objectives aimed at falsifying the requirements of the system under test.

Given a simulator $Sim$, an autonomous system under test (SUT) operates in a simulated environment $E$ defined within $Sim$. A simulated environment is defined in terms of temporal parameters $P$, static and dynamic objects $O_s$ and $O_d$. Examples of parameters $P$ include the illumination level, time of the day, weather conditions, etc. The value of these parameters may evolve during the scenario execution or remain static. Static and dynamic objects are defined via parameters $P_{os}$ and $P_{od}$, respectively, describing their location, orientation, shape, etc. For static objects, the values of their parameters $P_{os}$ remain constant during the simulation execution. For the dynamic objects, the values of $P_{od}$  evolve at each timestamp $t$ of running the simulation. Constraints $C$ can be imposed on the values describing $O_d$ and $O_s$. The SUT is defined by its state, $S$, that evolves at each timestamp $t$. In a given environment, the SUT has a mission $M$ to complete. 
Examples of missions include following a specific trajectory or navigating from a \textit{start} location to a \textit{goal} location. 
We define the environment \( E \) with the SUT 
assigned a specific mission \( M \) as a test scenario \( TS \). Upon executing \( TS \) in a simulator \( Sim \), the observed SUT states \( S_t \) are recorded in a system state trace \( T  = (S_0, t_0), (S_1, t_1), ..., (S_N, t_N)\), where $N$ is the number of simulation steps.
At all times of operation, SUT is expected to satisfy certain requirements \(R_i \in \mathbf{R} \), where $\mathbf{R}$ is the full set of requirements. Each requirement $R_i$ is specified via a corresponding signal temporal logic (STL) formula \( \varphi_i \). STL is a commonly used formalism for describing the behavior of cyber-physical systems~\cite{donze2010robust}. In our study, we consider the STL requirements in the following form:
\begin{align}
    \varphi_i = \square (s_i(t) > \epsilon_i)
\end{align}
which states that at each time step, i.e., the value of the system state signal \(s_i(t) \in T\) should ``always'' be greater than a certain threshold \(\epsilon\). A system state signal \( s_i \) represents specific values derived from the system state \( S \) that are used to assess whether the system meets a particular requirement \( R_i \). For example, in the simulation of an autonomous vehicle, relevant state values in \( S \) may include the deviation from the lane and the minimum distance to pedestrians. The state signal \( s_1 \), associated with the ``staying within the lane'' requirement, would correspond to the vehicle's deviation from the lane. Similarly, the state signal \( s_2 \), associated with the ''keeping a safe distance from pedestrians'' requirement, would represent the minimum distance to pedestrians.
We employ the 
robustness metric $\rho$~\cite{fainekos2009robustness} to evaluate the extent to which the property $\varphi_i$ is satisfied. Robustness $\rho_i$ of satisfying requirement $R_i$ typically corresponds to the difference between the minimum $s_i(t)$ observed and the established $\epsilon_i$ threshold: 
\begin{align}\label{eq:trace_deviation}
\rho_i = \min_{t} s_i(t) - \epsilon_i\
\end{align}
For instance, in the case of an AV avoiding collisions with pedestrians, $\rho_i$ can correspond to the difference between the minimum distance observed and the established safe distance threshold. If the AV maintains a safe distance, the value of $\rho_i$ will be higher.
When the SUT has multiple requirements $R_i$, we use the 
global robustness metric $\rho_{safe}$, to measure the extent to which the simulation trace $T$ falsifies the established requirements:
\begin{align}
    \rho_{safe}(T) = \min_{s_i \in T} \rho_{i}(s_i)
\end{align}
Robustness $\rho_{safe}$ corresponds to the lowest robustness $\rho_i$ observed in a trace T.
The goal of test generation is to manipulate the parameters of $TS$ in such a way that system traces  $\hat{T} $ with the lowest possible values of $\rho_{safe}$ are produced:
\begin{align}
    \hat{T} = \arg \min_{T \in \mathcal{L}(\Sigma)} \rho_{safe}(T)
\end{align}
were $\mathcal{L}(\Sigma)$ represents the set of all possible system traces. In order to solve this non-linear, non-convex optimization problem, stochastic search optimization methods can be applied~\cite{tuncali2018simulation}. In the context of testing ARS this is a challenging task. First, the possible search space for test scenarios is vast. Second, the produced traces $\hat{T}$ should be diverse. Third, running a simulator for collecting traces $T$ is often time-consuming due to the intensive computations required for scene rendering. In our work, we aim to facilitate the search by improving the representation of the problem.
In the following subsection, we describe the test generation problems for each of the two studied use cases.

\subsection{Obstacle positioning}
\begin{figure}[h!]
\begin{subfigure}{0.45\textwidth}
  \centering
  \includegraphics[scale=0.3]{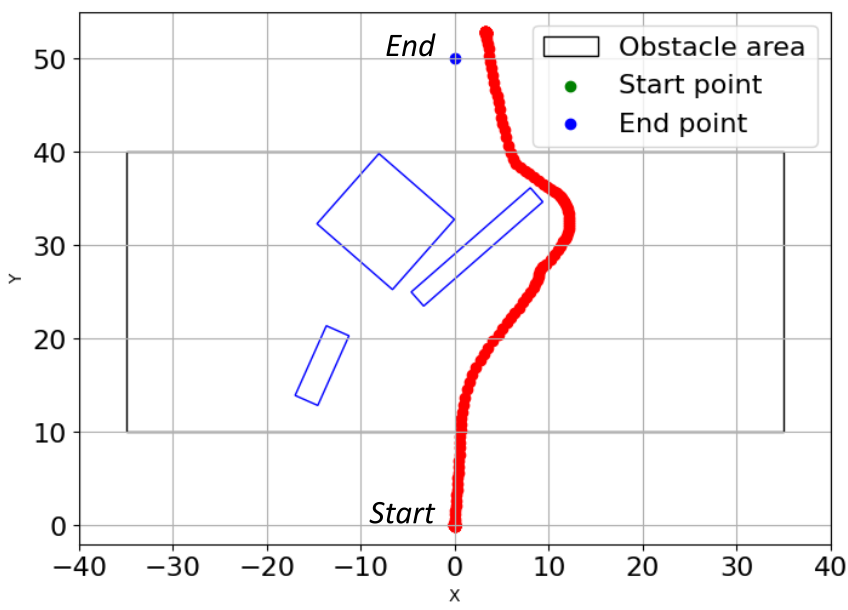}  
  \caption{Test example for UAV}
  \label{fig:uav_scenario}
\end{subfigure}
\begin{subfigure}{0.4\textwidth}
  \centering
  \includegraphics[scale=0.3]{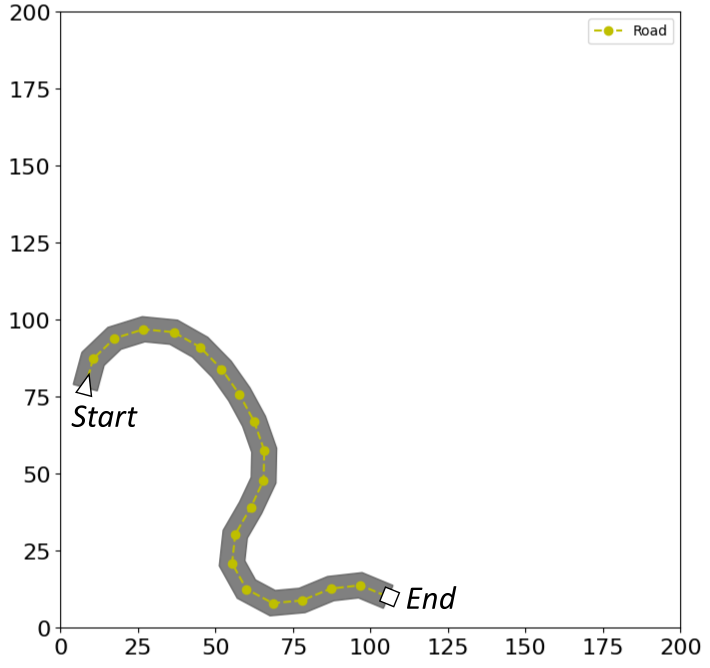}  
  \caption{Test example for UAV}
  \label{fig:ads_scenario}
\end{subfigure}
\caption{Examples of the considered test scenarios}
\label{fig:test_scenario}
\end{figure}
The first test generating problem we consider is the placement of rectangular shaped obstacles in an enclosed environment. This test generation problem was previously proposed by Khatiri et al. \cite{khatiri2024simulation} and used for the SBFT-2024 CPS-UAV competition \cite{khatiri2024sbft}. Such tests can be used to evaluate the performance of the navigation algorithm of an autonomous drone. Here, the temporal parameters $P$ such as weather and illumination correspond to normal conditions, with a sunny day illumination level and remain fixed. In this environment, there are no dynamic objects $O_d$; i.e., all objects are assumed to be static. The number of static objects $O_s$ varies from 1 to 3. Objects $O_s$ are rectangular obstacles, each described with 6 parameters, such as: $x$, $y$ coordinates, width ($w$), length ($l$), height($h$), and orientation ($r$). The following constraints $C$ are imposed on static objects: (1) objects cannot intersect, and (2) objects cannot completely block the path for the SUT. The mission of the SUT is to complete the path from the start to the goal location. The SUT has only one requirement, $R$, which is to remain at a distance of more than 1.5 m from the obstacles during the flight. Here, the system state signal that we record during the simulation execution is the minimal distance to any of the obstacles, which should be bigger than the threshold $\epsilon$ = 1.5. The goal of the test generation algorithm is to find such combinations of obstacles positions, shapes, and orientations that force the SUT to violate the requirement. An example of one test is illustrated in Figure \ref{fig:uav_scenario}. It consists of 3 obstacles with the drone trajectory shown in red. The mission consists of flying straight from the start to the goal location. The average execution time for a single test scenario ranges from 4 to 5 minutes. 

\subsection{Road topology generation}
In the second problem, the goal is to create road topologies that cause the lane keeping assist system (LKAS) of an autonomous vehicle to fail. This test generation problem was used at SBFT CPS testing competition 2023 \cite{biagiola2023sbft}. Similarly to the previous problem, the temporal parameters, such as weather and illumination, remain fixed and correspond to a sunny day. No dynamic objects $O_d$ are present in the environment. One static object $O_s$ is present, which is the road topology with two lanes. The road topology is defined with 2D points, that are then approximated with cubic splines to obtain a smooth trajectory. The number of road points can be variable, with at minimum two points, so that they can be approximated with splines. The following constraints $C$ are imposed on the road topology: (1) the road has to remain within the map borders, (2) the road cannot be too sharp, and (3) the road cannot self-intersect. An example of a road topology containing 17 2D control points is shown in Figure \ref{fig:ads_scenario}. The mission $M$ of the SUT is to follow the trajectory of the road topology from the start to the end location. These locations are given by the first and last points defining the road topology. The main requirement $R$ for the SUT is to stay within the lane boundaries, with a threshold of $p=0.85$ (as defined at the SBST 2021 competition~\cite{panichella2021sbst}), meaning that the vehicle is considered to be out of the lane if more than 85\% of it lies outside the lane boundaries. As the system state signal, we record the percentage of the vehicle that stays within the lane, which should be bigger than the threshold $\epsilon=0.15$.
The test generation algorithm should produce such road topologies that force the SUT to violate the requirements and go out of the lane bounds. One test scenario simulation takes from 0.5 to 1 minute on average.

\section{RILaST approach}\label{sec:RILaST}
\begin{figure}[h!]
\includegraphics[scale=0.55]{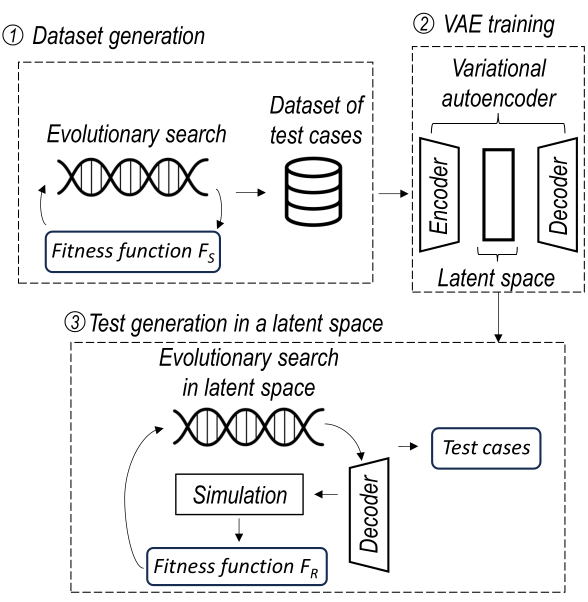}
\centering
\caption{Collecting a dataset and training a VAE}
\label{fig:step1}
\end{figure}

RILaST approach aims at improving the original test scenario representation. The new test representation obtained with RILaST should make it easier for the search algorithm to navigate the search space and converge towards failure revealing test cases.
The RILaST approach includes three key steps: (1) collecting a dataset of test scenarios, (2) training a variational autoencoder (VAE) model, and (3) executing an evolutionary search within the latent space of the autoencoder. These steps are illustrated in Figure \ref{fig:step1}. We present the detailed explanation of each of the steps below.
\subsection{Dataset generation}
An important step in VAE training is the collection of the dataset $D$. Inherently, a VAE learns to generate samples from a similar distribution as the training dataset. Therefore, it is important to ensure the dataset of test scenarios is diverse, but at the same time contains relevant and challenging test cases. 
Previous approaches in evolutionary search literature propose running a genetic algorithm~\cite{bentley2023using} or illumination search~\cite{gaier2020discovering} to produce an initial dataset for VAE training.
In the test generation context, we propose to first run a genetic algorithm guided by a simplified objective function several times, adding the population from the last generation to the dataset. We provide a more detailed description of the simplified function in the next subsection. We chose genetic algorithms as they have previously proven effective in test generation \cite{humeniuk2023ambiegen}. As another baseline, we also consider generating an entire dataset randomly, where each test is produced by uniformly sampling the allowed parameter ranges.
Running the search algorithm multiple times allows exploring a bigger search space, ensuring the diversity of the solutions in the dataset. Moreover, we add diversity preserving mechanisms to the search itself, to ensure that the solutions in the last generation are diverse. To estimate the number of runs $N$ of the genetic algorithm we divide the final dataset size $N_{data}$ by the population size, i.e., \(N = \frac{N_{data}}{P}\)
For both use cases, we collected datasets of 10 000 entries for running a genetic algorithm with a population of 200 for 50 generations 50 times.

\subsubsection{Simplified function design}\label{sec:surrogate_function}
\begin{figure}[h!]
\includegraphics[scale=0.5]{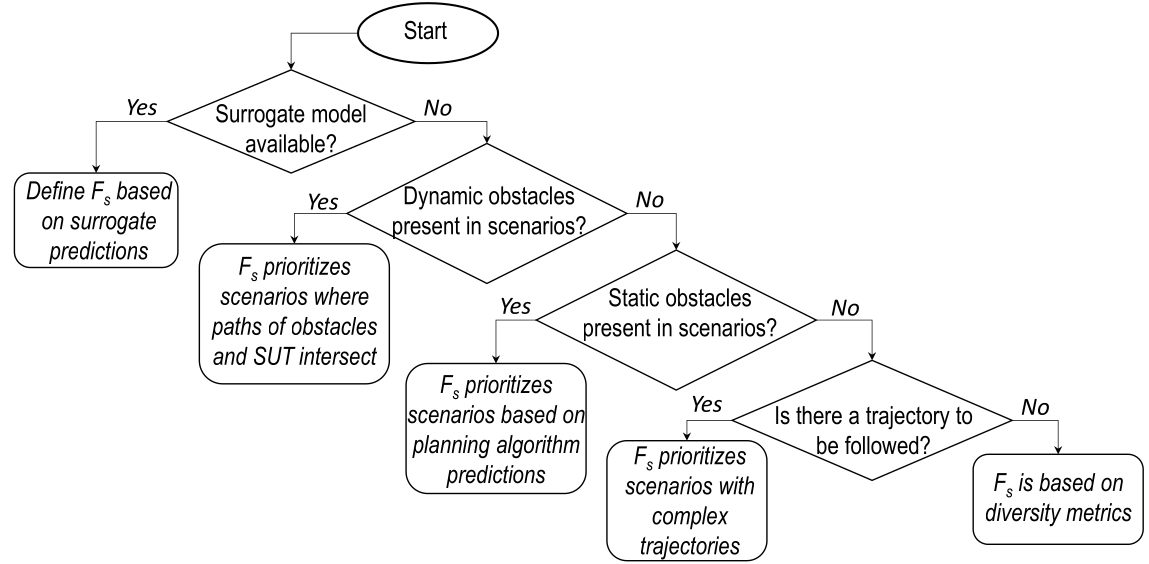}
\centering
\caption{Simplified fitness function design}
\label{fig:surrogate_function}
\end{figure}

To guide the generation, we propose using a simplified fitness function $F_s$ that is based on some knowledge of what constitutes a challenging test scenario. The function $F_s$ should be inexpensive to evaluate to be able to generate the dataset in a reasonable amount of time. In Figure \ref{fig:surrogate_function} we provide some guidance in choosing the function. 
Firstly, the knowledge can be contained in a simplified, i.e., surrogate model of the system. Such model is much less expensive to execute, and it provides approximate, but accurate enough results. Such models can be derived using traditional system identification methods \cite{menghi2020approximation} or supervised machine learning algorithms \cite{birchler2022cost}. For instance, Haq et al. \cite{haq2022efficient} developed a surrogate model for the Pylot \cite{gog2021pylot} system in the CARLA simulator. Birchler et al. \cite{birchler2022cost} introduced SDC-scissor, which is a supervised learning-based model predicting the output of the scenarios for LKAS testing. 
Moreover, some systems, such as mobile robots, have kinematic models available \cite{campion1996structural}, which can also serve to approximate their behavior. The surrogate model's predictions on the test scenario outcome can then be utilized as a fitness function during the collection of the dataset.

Next, if the surrogate model is not available, we suggest some heuristics from autonomous system testing literature. For instance, when designing test scenarios with dynamic obstacles, such as vehicles or pedestrians, Babikian et al. \cite{babikian2023concretization} propose to prioritize test cases where the paths of the SUT and the obstacles intersect. If static obstacles are present in the scenario, Humeniuk et al. \cite{humeniuk2022search} propose to prioritize test scenarios where the path planning algorithms, such as RRT or A* return a longer path. When the SUT is required to follow a certain trajectory, Zohdinasab \cite{zohdinasab2024focused} showed in their experiments that complex trajectories, involving turns of a high level of curvature, are more likely to cause failures. 
Finally, if none of these cases apply, test scenarios can be prioritized based on their novelty and the satisfaction of constraints. 
Specifically, we propose to run the novelty search algorithm, as proposed by Lehman et al. \cite{lehman2008exploiting}, where the objective function is defined as the average diversity between a given solution and its $n$ nearest neighbours.

Below we provide more details on the genetic algorithm configuration for the generation of the dataset.
When there are multiple falsification objectives, algorithms for multi-objective optimization such as NSGA-II \cite{deb2002fast} or MOEAD \cite{zhang2007moea} can be used. In our case, we leveraged a GA with a single optimization objective \cite{eiben2015introduction}.

\subsubsection{Genetic algorithm design} \label{sec:ga_design}
In the following, we present the implementation details of the genetic algorithm. We refer to this GA configuration as $GA_D$, in other words GA for dataset collection.

\textit{Individual representation and sampling.} We represent the individuals, or chromosomes, as one-dimensional arrays of size $M$. It is a standard way of representing individuals in evolutionary computation \cite{eiben2015introduction}, i.e., arrays of real values,  and it simplifies the use of existing search operators, such as crossover and mutations. 
For the obstacle placement use case (UC1), each test case is described by a 19-dimensional array. 
The first element corresponds to the number $n$ of box-shaped obstacles in the scene. The remaining 18 elements describe the obstacles that are present in the scene (up to 3). Each obstacle is described by 6 elements: x and y coordinates of the obstacle center position, obstacle length $l$, width $w$, height $h$ in meters, and rotation $r$ in degrees. 

\begin{table}[h!]
    \caption{Allowed ranges for the parameters for the studied use cases}
    \centering
    \scalebox{0.9}{
    \begin{tabular}{ccccccc|cc}
    \hline
      $n$ & $x$ & $y$ & $l$ &$w$  & $h$ & $r$  &$gb$ & $rb$ \\
    \hline
      $1-4$ & $-40-30$ & $10-40$ & $2-20$ &$2-20$  & $15-25$ & $0-90$ & $-0.07-0.07$ & $0.05$ \\
      \hline
    \end{tabular}}

    \label{tab:repr}
\end{table}
 For the second use case (UC2), to define the road topology points we leveraged a curvature-based representation introduced by Castellano et al. \cite{castellano2021analysis}, where a road topology is described by a sequence of $N$ curvature values (kappa) $\kappa_0$, $\kappa_1$, ..., $\kappa_{N-1}$. These values can then be mapped to 2D points with a simple transformation. Based on the experimental results, we chose to have a fixed size of a kappa vector of 17. Having a fixed-size vectors also simplifies the VAE training.
When generating the initial vectors of kappa values, the value of $\kappa_i$ is obtained from a range
defined by a global bound and a relative bound. The global
bound ($gb_l$, $gb_u$) defines the minimum and the maximum
value that $kappa_i$ can take, while the relative bound $rb$ defines
the difference between $k_i$ and its previous value. The values of these bounds are indicated in Table \ref{tab:repr}.
In both use cases, for the initial population, we generate $N_{pop}$ vectors randomly by sampling from the allowed value ranges, which are listed in Table \ref{tab:repr}. $N_{pop}$ corresponds to the population size.

\textit{Fitness function.} 
To guide the search for dataset collection, we used a simplified $F_s$ fitness function. In the obstacle placement problem, the goal is to find a configuration of static objects that will reveal a SUT failure. Based on Figure \ref{fig:surrogate_function} we select $F_s$ to be based on the RRT* planning algorithm \cite{lavalle1998rapidly}.  Intuitively, the longer the path produced by the planner, the more deviations from the shortest path the UAV will have to make, increasing the risk of its failure.

For the topology generation, we used the maximum road curvature value as the objective function. Intuitively, the bigger the curvature angle is present in the road topology, the more challenging it should be for the driving agent.  When the maximum road curvature exceeded the limits, the test cases are given a low fitness value.

\textit{Crossover operator.} We are using a one-point crossover operator, which is one of the commonly-used operators \cite{umbarkar2015crossover}. Given a pair of parents, the crossover point is randomly determined. At this point, genes 
(components of the chromosomes) are exchanged. This operator is applied with probability $p_{cross}$. Following the generation of the offspring, test scenario constraint $C$ feasibility validation should be performed. If the constraint is violated, a low fitness value is assigned to the individual. 

\textit{Mutation operator.} For both use cases, we use domain-specific mutation operators. After applying the crossover operator, one of the presented operators is applied with a probability $p_{mut}$.
For the obstacle placement problem, three operators are used: random modification to one obstacle, obstacle addition, or obstacle removal. During obstacle modification, one of the obstacles was randomly selected and its parameters, such as $x, y, w, h, l, r$ were assigned random values from the available ranges. During obstacle addition, one obstacle with randomly sampled parameters is added, if the maximum number of obstacles in the scene is not exceeded. During the obstacle removal, one obstacle is randomly selected and removed, ensuring that at least 1 obstacle is left in the scene. 

For the road topology generation, we leveraged the mutation operators proposed by the FreneticLib framework \cite{klikovits2023frenetic}. These operators include: increasing kappa values from 10 \% to 20 \% (increase operator), changing the values of $n$ randomly selected kappas (change operator), reversing the order of kappa values (reverse operator), changing the sign of kappa values (sign change operator). In `increase' operator, the value of the increase is uniformly sampled from 10 \% and 20 \%  and applied to all the kappa values. In `change' operator, we randomly select up to 5 kappa values and randomly change their values within global bounds $gb$. In the reverse operator, we reverse the order of all the kappa values. In `sign' change operator, the signs of all the kappa values are changed to the opposite one. Having a diverse set of mutations helps generate a set of diverse test scenarios.

\textit{Duplicate removal.} At each iteration, we remove the individuals which have a small deviation between them. We estimate the deviation as the cosine distance $D_c$ \cite{foreman2014cosine} and compare it to a prefixed minimum threshold, $D_{cth}$. If the $D_c$ between the two individuals is less than $D_{cth}$, one of them is removed. 

The hyperparameters used for this GA configuration are specified in Table \ref{tab:hyper_param_gad}.
\begin{table}[h]
\caption{Hyperparameters of GA configuration for the dataset collection}
\label{tab:hyper_param_gad}
\centering
\begin{tabular}{cccccc}
\hline
Parameter     & Description & Values   \\
\hline
$N_{pop}$     & Population size & 200 \\
$p_{cross}$        & Crossover rate      & 0.9  \\
$\eta_{c}$        & Initial crossover $\eta$      & 3  \\
$p_{mut}$         & Mutation probability       & 0.4    \\ 
$\eta_{m}$         & Initial mutation $\eta$       & 3    \\   
$D_{cth}$         & Duplicate threshold       & 0.025    \\
  $N_{off}$         & Number of offspring       & 100  \\

\hline
\end{tabular}
\end{table}

\subsection{VAE training}
\begin{figure}[h!]
\includegraphics[scale=0.6]{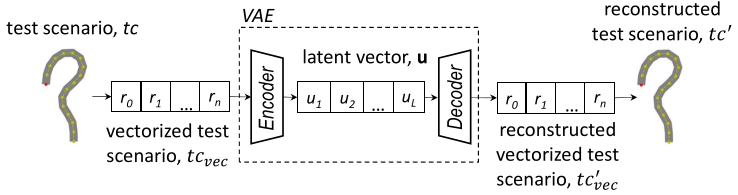}
\centering
\caption{Test case encoding and decoding with VAE}
\label{fig:vae_diag}
\end{figure}
The primary objective of training the VAE is to learn a mapping function between a given test case and its representation in a latent space. The process of transforming a test to the latent space and then back into the original space is shown in Figure \ref{fig:vae_diag}. Test scenarios $tc$ are given as the input to the simulation environment. 
In the figure, the test corresponds to a road topology, represented as a sequence of 2D points. As presented earlier, in order to perform optimization, test scenarios should be transformed to a simpler genotype representation, such as 1D array of size $M$, which we refer to as vectorized test scenario $tc_{vec}$. Mapping from $tc$ to $tc_{vec}$ typically is done with a deterministic function. Traditional search-based approaches perform the search in the vectorized test scenario space. Our goal is to use the obtained dataset $D$ and the VAE to learn a better representation to perform the search in. We refer to this representation as the latent vector $u$. The search space of latent vectors is referred to as latent space. To map the $tc_{vec}$ to $\textbf{u}$, the encoder part $M_E$ of the VAE is used. At the output of the encoder we get the latent vector $\textbf{u}$ of size $L$. The values of this vector follow a normal distribution with a mean of 0 and a standard deviation of 1. The size of $L$ is typically smaller or equal to $M$. Reducing the size of $L$ can help reduce the search space size, but at the same time it can increase the reconstruction error of the original input $tc_{vec}$. Therefore, we chose to have the size of $L$ equal to the input space size $M$. To map the vector from the latent space $u$ to the original vectorized test scenario space, the decoder module $M_D$ is used. Passing $\textbf{u}$ through the decoder, we obtain the reconstructed vectorized test scenario $tc_{vec}^{'}$. With a trained VAE, we expect the mean squared error (MSE), i.e., reconstruction loss between the $tc_{vec}$ and $tc_{vec}^{'}$ to be minimized. Achieving low reconstruction loss is important to ensure that the VAE learns a meaningful representation. A poorly learned latent representation results in an inconsistent mapping between the latent variable and the actual variable, complicating the evolution process \cite{bentley2022coil}. Having a well reconstructed $tc_{vec}^{'}$ is also important for obtaining meaningful and valid reconstructed test scenarios $tc^{'}$ in the end. Below, we provide more details on our VAE training setup.

VAE is a neural network, consisting of two main parts: encoder and decoder, with symmetric architecture. During the training phase, both encoder and decoder parts are used. During the search in the latent space only the decoder is required. We trained the VAE with a set-up recommended by Bentley et al. \cite{bentley2022coil}. The encoder consists of two hidden layers with sizes 128 and 64, respectively. The decoder has the same architecture but in reverse order. The VAE was trained for 1000 epochs using the Adam optimizer with a learning rate of 0.001 and a batch size of 512. To ensure smoother activation values during training, we normalized the dataset inputs to the range $[-1, 1]$ and applied the $\tanh$ function at the output layer. The VAE learns to represent the data within a latent space $\mathbf{z}$, where latent vectors $z_i \in \mathbf{z}$.  To minimize the final loss, we set the dimensionality of the latent vectors equal to the input size.
We train a separate VAE for each use case, which we refer to as $VAE_{uc1}$ and $VAE_{uc2}$ for the first and second use cases, respectively. 

\subsection{Test generation in the latent space} \label{sec:latent_space}
Having an improved representation in a latent space of the VAE, we can perform the search over this space. We leverage a genetic algorithm with a simulator-based objective function. We also use the decoder part of the VAE $M_D$ to map the test specification in the latent space to the original space. Below, we provide a more detailed description of the GA we use to perform the search. We refer to this GA configuration as $GA_L$, in other words, GA for optimization in a latent space.

\textit{Individual representation and sampling.}  We represent the individuals, or chromosomes, as one-dimensional arrays of the size of the latent space $L$. In our implementation, the latent space size is equal to the original representation size. 
For the obstacle placement use case, each test case is described by a 19-dimensional latent vector, while for the topology generating use case, a test case is described by a 17-D vector. 
The key feature of the latent space is that each dimension of the latent vector $\mathbf{u}$ is designed to follow a normal distribution with a mean of 0 and a standard deviation of 1, i.e., $\mathbf{u} \sim \mathcal{N}(0, I)$, where $I$ is the identity matrix. Therefore, to generate the random population, we simply sample $L$ values from a normal distribution $\mathcal{N}(0, 1)$ to create each of the $N_{pop}$ latent vectors $\mathbf{u}$. Here, $N_{pop}$ corresponds to the population size.

\textit{Fitness function.} 
To guide the search in the latent space, we use the simulator-based fitness function $F_{\text{sim}}$.
We evaluate it based on Eq. (\ref{eq:trace_deviation}), which computes the deviation between the recorded system execution trace $T$, i.e., the observed behavior, and the given requirement, i.e., the expected behavior. The system execution trace is obtained after running the given test scenario in the simulation environment. For UC1, as the system trace $T$, we recorded the closest distance from the SUT to an obstacle at each timestamp. The SUT requirement $R$ was to complete the flight mission without approaching the obstacle too closely, i.e., flying at a distance of more than 1.5 meters. As the fitness function $F_{\text{sim}}$, we calculated the minimal distance to the object observed. For UC2, we recorded the percentage of the SUT going out of the road bounds. The requirement was to avoid going out of the road bounds. As the fitness function, we calculated the maximum percentage of the SUT observed to be out of the lane.

\textit{Crossover operator.} The representation in the latent space, i.e., a vector of $L$ independent real values, aligns with the standard representation used in the evolutionary search domain \cite{eiben2015introduction}. Therefore, all real-valued search operators developed by the evolutionary search community can be applied. We selected the self-adaptive simulated binary crossover \cite{deb2001self} as one of the most advanced search operators. Simulated binary crossover exchanges the values between two parents based on the distribution index $\eta_c$, where a small value encourages more exploration (offspring are further from parents), while a large value encourages more exploitation (offspring are close to parents). Self-adaptation allows the evolutionary algorithm to dynamically adjust the distribution index ($\eta_c$) during the search process, based on the offspring performance. 

\textit{Mutation operator.} After the crossover operator, each chromosome undergoes a polynomial mutation operator \cite{deb2007self} with $p_{mut}$. For each gene (variable) in the chromosome, a perturbation is introduced based on a polynomial distribution controlled by the mutation distribution index $\eta_m$. Larger values of $\eta_m$ generate offspring closer to the original. This perturbation is applied with probability $\frac{1}{n_{var}}$, where $n_{var}$ corresponds to the number of variables in the chromosome.

\textit{Duplicate removal.} At each iteration, we remove the individuals that have a small deviation between them. We estimate the deviation as the cosine distance $D_c$ \cite{foreman2014cosine} and compare it to a prefixed minimum threshold, $D_{cth}$. If the $D_c$ between the two individuals is less than $D_{cth}$, one of them is removed. 

The hyperparameters used for this GA configuration are specified in Table \ref{tab:hyper_param_latent}. We used the same hyperparameters for both test generation problems.
\begin{table}[h]
\caption{Hyperparameters of GA configuration for running RILaST}
\label{tab:hyper_param_latent}
\centering
\begin{tabular}{cccccc}
\hline
Parameter     & Description & Values   \\
\hline
$N_{pop}$     & Population size & 40 \\
$p_{cross}$        & Crossover rate      & 0.5  \\
$p_{mut}$         & Mutation probability       & 0.4    \\   $D_{cth}$         & Duplicate threshold       & 0.025    \\
  $N_{off}$         & Number of offspring       & 20  \\

\hline
\end{tabular}
\end{table}

We present the overall process of the optimization in the latent space in Algorithm \ref{alg:RILaST}.
Like every evolutionary algorithm, RILaST starts by creating an initial population $P$ of size $N_{\text{pop}}$. Differently from existing initial population sampling techniques, in RILaST the population is created by randomly sampling L-dimensional vectors with values from a normal distribution. Then, each of the vectors in the population is assigned a fitness value based on the evaluation in the simulated environment. To perform the evaluation, each latent vector $u$ is first decoded with the decoder model $M_D$ into the vectorized test scenario. We then use a transformation function $\phi_{\text{tc}}$ to map the test scenario from the vectorized form to the test scenario specification $tc$ for the simulated environment. We run the simulation and compute the value of the objective function $F_{\text{sim}}$. In each of our case studies, we only have one objective function, but multiple objectives can be added if needed.

When solutions are evaluated, tournament selection is used to choose the individuals for crossover and mutation, which are then performed on individuals with probabilities $p_{\text{cross}}$ and $p_{\text{mut}}$, respectively. The obtained offspring $P_{\text{offspring}}$ are evaluated using the same process as at the beginning of the search. We then use an elitist approach to select individuals for the next generation, where the $N_{\text{pop}}$ individuals with the highest $F_{\text{sim}}$ values are selected from the merged population of $P$ and $P_{\text{offspring}}$. The search continues until a termination criterion is met. In our case, we used the number of evaluations or the running time as the termination criteria.

\begin{algorithm}
\caption{AmbieGenVAE: Test generation algorithm in the latent space}\label{alg:RILaST}
\begin{algorithmic}[1]
\Require Weights of the pre-trained VAE decoder $M_D$
\Require Simulator based evaluation function $\phi_{sim}$
\Require Test scenario transformation function $\phi_{tc}$
\Require Termination criterion $T$
\Require Hyperparameters $N_{\text{pop}}, N_{\text{off}}, \text{$p_{cross}$}, \text{$p_{mut}$}$
\State Set hyperparameters, initialize population $P$.
\State Load pre-trained VAE model
\State Initialize population $P$ with $N_{pop}$ individuals
\State Evaluate initial population:
\For{$n = 1$ to $N_{\text{pop}}$}
    \State Decode latent vector $P_n$ into vectorized test scenario: $tc_{vec} \leftarrow M_D(P_n)$
    \State Convert vectorized test to original representation: $tc \leftarrow \phi_{tc}(tc_{vec})$
    \State Evaluate fitness: $F_{sim_n} \leftarrow \phi_{sim}(tc)$
\EndFor

\While{not $T$}
    \State Select individuals for reproduction: $P_{\text{selected}} \leftarrow \text{TournamentSelection}(P)$
    \State Perform crossover: $P_{\text{crossover}} \leftarrow \text{Crossover}(P_{\text{selected}}, p_{\text{cross}})$
    \State Perform mutation: $P_{\text{offsring}} \leftarrow \text{Mutation}(P_{\text{crossover}}, p_{\text{mut}})$
    \State Evaluate offspring (as in initial population evaluation):
    \For{$n = 1$ to $N_{\text{off}}$}
        \State $tc_{vec} \leftarrow M_D(P_n)$
        \State $tc \leftarrow \phi_{tc}(tc_{vec})$
        \State $F_{sim_n} \leftarrow \phi_{sim}(tc)$
    \EndFor
    \State Select survivors: $P \leftarrow \text{SurvivorSelection}(P, P_{\text{offspring}})$
\EndWhile
\State \Return final population $P$

\end{algorithmic}
\end{algorithm}

\section{Evaluation}\label{sec:evaluation}

In this section, we formulate and answer the research questions related to the performance of RILaST. Our primary objective is to assess the extent to which the search with representation in the latent space of a VAE enhances the performance of test generators. 
To compare the algorithms, we assessed the following metrics: the average number of revealed failures ($F_{av}$) and their diversity.
For the obstacle generation problem, we classify the test as failed if the SUT approaches the obstacles at a distance less than 1.5 meters. For the topology generation problem, a test is considered failed if more than 85 \% of the SUT goes out of the lane bounds.

We evaluated the diversity of test scenarios based on the diversity of input test specifications. We did not consider the output diversity in terms of system behavior, as this metric is influenced by the stochasticity of the simulator. Intuitively, diverse input environment specifications should lead to diverse system behaviors. 
To evaluate the diversity, we leverage the failure sparseness, $S$, a metric proposed in the literature \cite{riccio2020model}. It estimates the diversity by computing the average value of the distance metric between each pair of the failed test cases $tc$:
\begin{equation}\label{eq:spars}
S = \frac{\sum_{i}^{N} ave_{j}^{N}\text{dist}(tc_i, tc_j)}{N}
\end{equation}
where $N$ is the total number of failed test cases, $ave$ represents the average value, and $\text{dist}$ is the distance metric.
In UC1, we used the cosine distance \cite{foreman2014cosine} as the distance metric $dist$. For UC2, we calculated $dist$ as the weighted Levenshtein distance~\cite{levenshtein1966binary}
between road segments near the location where the failure happened (car went out of the bounds) as suggested by Riccio and Tonella~\cite{riccio2020model}. This metric was also used at the SBST 2021 tool competition~\cite{panichella2021sbst}.

\subsection{Research questions}

We formulate the following research questions:
\begin{enumerate}
 \item \textbf{RQ1} (Effectiveness of RILaST compared to baselines).
\textit{To what extent does RILaST improve the identification of diverse failures in autonomous robotic systems compared to baseline approaches?}
In $RQ1$ we compare RILaST performance in terms of number of failures found and their diversity 
to baseline approaches such as random search (RS) and genetic algorithm in original space as well as known approaches from the literature, such as AmbieGen \cite{humeniuk2023ambiegen}, TUMB~\cite{tang2024tumb}, RIGAA~\cite{rigaa} and CRAG~\cite{arcaini2023crag}.
\item \textbf{RQ2} (Effectiveness of search in latent space). \textit{To what extent does optimization in the latent space improve the identification of diverse failures compared to optimization with the original representation?} In this RQ, we compare the optimization in the original space with optimization in the latent space of two different VAEs: one trained with a randomly produced dataset and the other with an optimized dataset based on a simplified fitness function $F_s$. 
\item \textbf{RQ3} (Choice of VAE hyperparameters).
\textit{What VAE configuration leads to the best performance in terms of reconstruction loss?} In this RQ we evaluate 18 VAE configurations with three different architectures, learning rate, and batch size values. Our goal is to identify the hyperparameters that lead to better performance. This aspect has not been thoroughly explored in previous work.
\item \textbf{RQ4} (Impact of the number of latent variables on the VAE performance).
\textit{What is the impact of different VAE configurations in terms of the latent space size on the resulting validation loss and reconstruction fidelity?}
In this research question, we seek to explore how varying the number of latent dimensions of VAE during training impacts the model's performance. Specifically, we investigate how reducing as well as increasing the number of variables affects the final VAE loss as well as reconstruction fidelity.
\item \textbf{RQ5} (RILaST overhead). 
\textit{What is the overhead of RILaST in terms of dataset collection time, VAE training and inference?}
This research question focuses on evaluating the resource demands associated with training a VAE as part of the RILaST framework. We aim to quantify the overhead in three key areas: (1) the time required to collect and pre-process the dataset needed for VAE training, (2) the computational cost and time involved in training the VAE itself, and (3) the time taken for inference when generating scenarios in the latent space.
\end{enumerate}
\subsection{Systems under test}
In our study, we consider two simulator-based autonomous robotic systems under test: an autonomous unmanned aerial vehicle in the Gazebo simulator~\cite{1389727}  as well as an autonomous vehicle in the BeamNG simulator~\cite{beamng}.
\begin{figure}[h!]
\begin{subfigure}{0.45\textwidth}
  \centering
  \includegraphics[scale=0.37]{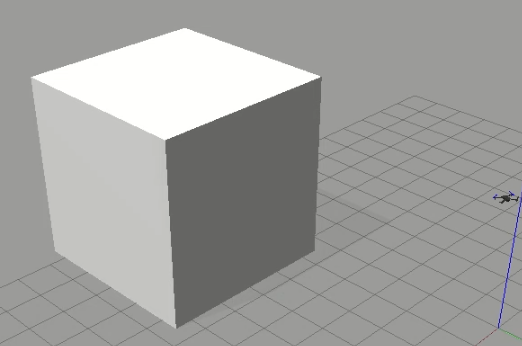}  
  \caption{Autonomous UAV in Gazebo simulator}
  \label{fig:rob_agent_rl}
\end{subfigure}
\begin{subfigure}{0.45\textwidth}
  \centering
  \includegraphics[scale=0.35]{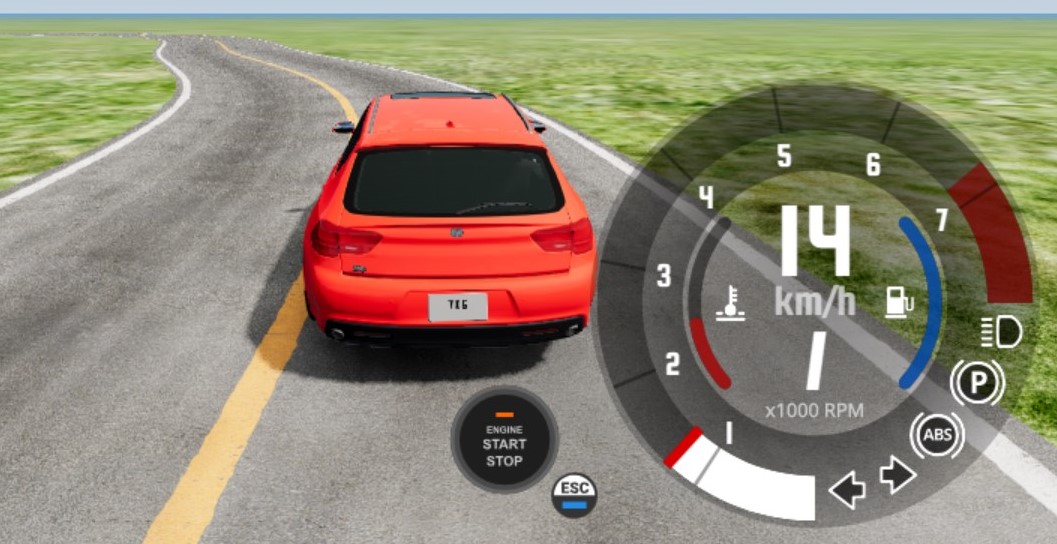}  
  \caption{Autonomous vehicle in BeamNG simulator}
  \label{fig:veh_beam}
\end{subfigure}
\caption{Autonomous UAV and autonomous vehicle systems used in our study}
\label{fig:agents}
\end{figure}

The autonomous UAV, shown in Figure~\ref{fig:rob_agent_rl} is an Iris 1
drone running PX4 2 control firmware \cite{meier2015px4}. This testing pipeline is based on the work of Khatiri et al. \cite{khatiri2023simulation, khatiri2024simulation}. The drone is tasked to navigate safely in a closed space with box-shaped obstacles. The drone is expected to follow a given trajectory autonomously, avoiding the obstacles. The goal of test generation is to reveal the scenarios where the UAV fails to complete the mission and approaches the obstacles at an unsafe distance of less than 1.5 m. This SUT corresponds to the first use case.

As the second test subject, we used an autonomous lane-keeping assist system agent BeamNG.AI, built-in in the BeamNG.tech~\cite{beamng} simulator, shown in Figure~\ref{fig:veh_beam}. The goal of the agent is to navigate a given road without going out of its bounds.
It has knowledge of the geometry of the entire road and uses a complex optimization process to plan trajectories that keep the ego-car as close as possible to the speed limit while staying inside the lane as much as possible. Test generation aims to find cases where the vehicle fails to follow the given trajectory. This SUT corresponds to the second use case we presented in Section \ref{sec:problem}.
\subsection{Results}
In this subsection, we present our experimental results for each RQ formulated above.

 \subsubsection{RQ1 – Effectiveness of RILaST compared to baselines}
\hfill
 
\textbf{Motivation.} In this research question, we aim to evaluate the ability of RILaST to generate diverse and challenging test scenarios. We compare RILaST to random search, GA in original space as well as to GA in original space guided by a combination of $F_s$ and $F_{sim}$, where $F_{sim}$ is used only for most promising tests according to $F_{s}$. This is an alternative way of embedding domain knowledge in the search process. We also compare RILaST with some of the known approaches from the cyber-physical system (CPS) testing literature.

\textbf{Method}. For both use cases, the baseline approaches include random search, a genetic algorithm, and the adapted AmbieGen method \cite{humeniuk2024ambiegen}, where a simplified function is used to guide the search. In random search, test scenarios are generated by sampling randomly from the allowed parameter ranges specified in Table \ref{tab:repr}. The GA baseline follows the $GA_L$ configuration for the search in latent space described in Section \ref{sec:ga_design}. However, instead of latent representation, it uses the same representation as GA for dataset collection $GA_D$. The next baseline, AmbieGen, uses both, simplified and simulator-based fitness functions to guide the search. In this method, $F_{sim}$ is applied only when the $F_s$ function predicts high fitness values, exceeding the established threshold. The simplified fitness function $F_s$ was configured identically to the one used for RILaST dataset collection GA configuration $GA_D$. Further details about this baseline can be found in our report \cite{humeniuk2024ambiegen} for the SBFT Tool Competition 2024 CPS-UAV track \cite{uav-competition}.
As the further baselines, for UC1, we used the available approaches for UAV test generation, such as 
 Tumb \cite{tang2024tumb} tool, which showed one of the best results at the CPS-UAV testing tool competition in 2024 \cite{uav-competition}. For UC2, we use the approaches for LKAS testing, namely RIGAA \cite{rigaa} and CRAG \cite{arcaini2024crag}, which are the finalists of the SBFT CPS-ADS testing tool competition in 2024 \cite{10.1145/3643659.3643932}.
 
We compare the performance of RILaST to baseline methods in terms of the average number of revealed failures ($F_{av}$), as well as the average sparseness ($S$) of the failures. For the UC1, we allocate a budget of 200 evaluations to all the compared approaches, as it was done in the CPS-UAV testing competition. For the UC2, we allocate a two-hour time budget for each run of the algorithms.  We run each baseline at least 10 times.

\begin{figure}[h!]
\begin{subfigure}{0.45\textwidth}
  \centering
\includegraphics[scale=0.37]{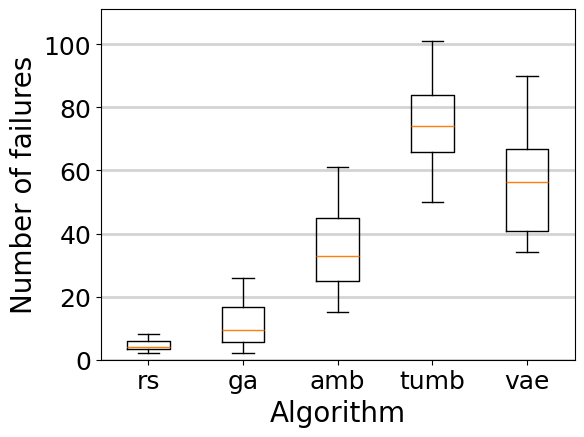} 
  \caption{The average number of failures for UAV (UC1)}
  \label{fig:rq1_uav_fail}
\end{subfigure}
\begin{subfigure}{0.45\textwidth}
  \centering
  \includegraphics[scale=0.37]{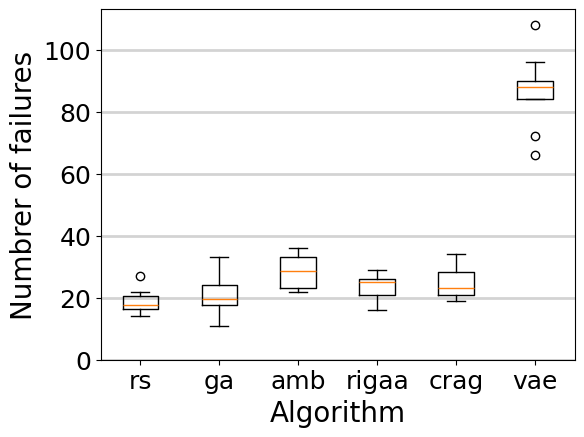}  
  \caption{The average number of failures for ADS (UC2)}
  \label{fig:rq1_ads_fail}
\end{subfigure}
\caption{The average number of failures detected by different algorithms for the simulator-based autonomous robotic systems}
\label{fig:rq1_fail}
\end{figure}

\begin{figure}[h!]
\begin{subfigure}{0.45\textwidth}
  \centering
\includegraphics[scale=0.37]{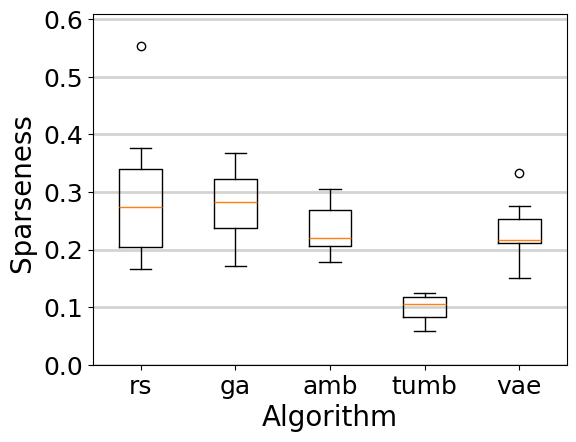} 
  \caption{The average sparseness of failures of the UAV}
  \label{fig:rq1_uav_in}
\end{subfigure}
\begin{subfigure}{0.45\textwidth}
  \centering
  \includegraphics[scale=0.37]{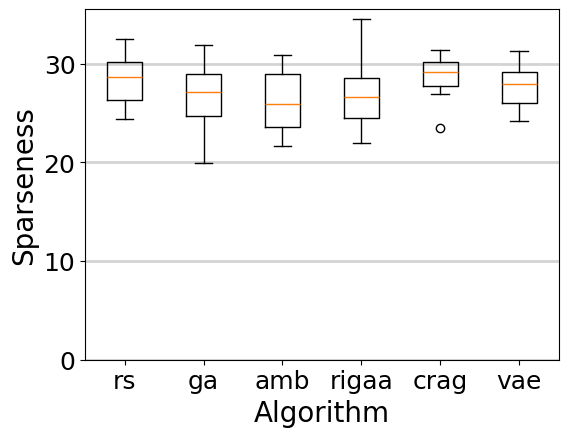}  
  \caption{The average sparseness of failures of the ADS}
  \label{fig:rq1_ads_in}
\end{subfigure}
\caption{The average sparseness of failures detected by different algorithms for the simulator-based autonomous robotic systems}
\label{fig:rq1_in_diversity}
\end{figure}

\textbf{Results.} The box plots for the average number of revealed failures and their sparseness for UC1 and UC2 are presented in Figure~\ref{fig:rq1_fail} and Figure~\ref{fig:rq1_in_diversity}, respectively. Tables~\ref{tab:rq1-uav2}–~\ref{tab:rq1-ads2} report the obtained mean values and statistical test results. In the plots, we refer to AmbieGen as `amb' and RILaST as `vae'. For the rest of the tools, the name corresponds to their acronym.

\begin{table}[!ht]
    \centering
    \caption{Number of revealed failures and their sparseness in uav case study}
    \scalebox{0.9}{
    \begin{tabular}{cccccccccc}
    \hline
        Metric & A & B & p-value & Effect size & Metric & A & B & p-value & Effect size \\ \hline
        Failures & rs & ga & 0.025 & -0.6, L & Sparseness & rs & ga & 0.97 & -0.02, N \\ 
        ~ & rs & amb & <0.001 & -1.0, L & ~ & rs & amb & 0.307 & 0.273, S \\ 
        ~ & rs & tumb & 0.001 & -1.0, L & ~ & rs & tumb & <0.001 & 1.0, L \\ 
        ~ & rs & vae & <0.001 & -1.0, L & ~ & rs & vae & 0.254 & 0.286, S \\ 
        ~ & ga & amb & 0.001 & -0.855, L & ~ & ga & amb & 0.084 & 0.455, M \\ 
        ~ & ga & tumb & 0.001 & -1.0, L & ~ & ga & tumb & <0.001 & 1.0, L \\ 
        ~ & ga & vae & <0.001 & -1.0, L & ~ & ga & vae & 0.05 & 0.486, L \\ 
        ~ & amb & tumb & 0.001 & -0.935, L & ~ & amb & tumb & <0.001 & 1.0, L \\ 
        ~ & amb & vae & 0.007 & -0.643, L & ~ & amb & vae & 0.805 & 0.065, N \\ 
        ~ & tumb & vae & 0.044 & 0.561, L & ~ & tumb & vae & <0.001 & -1.0, L \\ \hline
    \end{tabular}}
    \label{tab:rq1-uav2}
\end{table}
\begin{table}[!ht]
    \centering
    \caption{Number of revealed failures and their sparseness in ads case study}
    \scalebox{0.9}{
    \begin{tabular}{cccccccccc}
    \hline
        Mertic & A & B & p-value & Effect size & Mertic & A & B & p-value & Effect size \\ \hline
        Failures & rs & ga & 0.343 & -0.26, S & Sparseness & rs & ga & 0.345 & 0.26, S \\ 
        ~ & rs & amb & 0.001 & -0.89, L & ~ & rs & amb & 0.14 & 0.4, M \\ 
        ~ & rs & rigaa & 0.03 & -0.6, L & ~ & rs & rigaa & 0.391 & 0.244, S \\ 
        ~ & rs & crag & 0.004 & -0.725, L & ~ & rs & crag & 0.621 & -0.133, N \\ 
        ~ & rs & vae & <0.001 & -1.0, L & ~ & rs & vae & 0.473 & 0.2, S \\ 
        ~ & ga & amb & 0.011 & -0.68, L & ~ & ga & amb & 0.678 & 0.12, N \\ 
        ~ & ga & rigaa & 0.219 & -0.344, M & ~ & ga & rigaa & 1 & 0.0, N \\ 
        ~ & ga & crag & 0.074 & -0.458, M & ~ & ga & crag & 0.223 & -0.317, S \\ 
        ~ & ga & vae & <0.001 & -1.0, L & ~ & ga & vae & 0.623 & -0.14, N \\ 
        ~ & amb & rigaa & 0.12 & 0.433, M & ~ & amb & rigaa & 0.775 & -0.089, N \\ 
        ~ & amb & crag & 0.098 & 0.425, M & ~ & amb & crag & 0.07 & -0.467, M \\ 
        ~ & amb & vae & <0.001 & -1.0, L & ~ & amb & vae & 0.345 & -0.26, S \\ 
        ~ & rigaa & crag & 0.858 & -0.056, N & ~ & rigaa & crag & 0.166 & -0.37, M \\ 
        ~ & rigaa & vae & <0.001 & -1.0, L & ~ & rigaa & vae & 0.488 & -0.2, S \\ 
        ~ & crag & vae & <0.001 & -1.0, L & ~ & crag & vae & 0.199 & 0.333, M \\ \hline
    \end{tabular}}
    \label{tab:rq1-ads2}
\end{table}
For UC1, we can observe that RILaST outperforms both GA and RS in the number of revealed failures, identifying 4.64 times and 12.2 times more failures, respectively. While the sparseness of the test scenarios generated by RILaST is relatively high, it is, on average, 20\% lower than that of randomly generated scenarios. RILaST reveals 59\% more failures than the AmbieGen approach with no statistically significant difference in their diversity. Both approaches rely on the same simplified function to guide the test generation: RILaST during the dataset collection and AmbieGen during test execution. This result confirms that a smoother search space in the latent dimensions of the VAE positively influences the search process.
TUMB reveals 33.4 \% more failing test scenarios than RILaST. At the same time, the sparseness of the failing scenarios found by TUMB is 2.35 times lower than for the scenarios found by RILaST. 

For the ADS case study, RILaST generates, on average, from 3 to 4.7 times more failures than the rest of the tools. Specifically, it produces 3.68, 3.5, and 3 times more failures than the state-of-the-art tools RIGAA, CRAG, and AmbieGen, respectively. Meanwhile, all tools exhibit a high level of failure sparseness, with no statistically significant differences between them. On average, the CRAG tool achieves the highest sparseness value of 28.78.

\begin{tcolorbox}
\textbf{Summary of RQ1:} RILaST is effective in generating failure-revealing scenarios for autonomous robotic systems. In the UAV case study, it outperforms AmbieGen and GA by 59\% and 464\%, respectively. While RILaST generates 33\% fewer failures than TUMB, its scenarios are 2.35 times more diverse. In the ADS case study, RILaST generates 3 to 4.7 times more failures than other tools, including 3.5 times more than RIGAA and CRAG, while maintaining a high level of failure sparseness.
\end{tcolorbox}

\subsubsection{RQ2 – Effectiveness of search in latent space}
\hfill

\textbf{Motivation.} This research question evaluates the effectiveness of test generation in latent space compared to the original space. A key advantage of the latent space is its regularization, which provides a smoother search landscape. Additionally, in the latent space, tests are sampled from a distribution similar to the training data, incorporating domain knowledge into the test generation process. This biases the search toward higher-performing solutions and reduces the test budget spent on non-promising cases. Here, we aim to understand the impact of these benefits on test generation.

\textbf{Method.} To answer this RQ, we consider three different search spaces: the original space, the latent space of a VAE trained on randomly sampled tests, and the latent space of a VAE trained on tests previously optimized using a simplified fitness function.
In each search space, we compare the performance of random search (denoted as `ran' in the plots), genetic algorithm with problem-specific search operators (ga1), as well as the genetic algorithm with standard vector-based search operators (ga2). Random search samples the solutions uniformly from the allowed parameter ranges.
Genetic algorithms $ga1$ and $ga2$ follow the configuration presented in Table~\ref{tab:hyper_param_latent} with different search operators. Algorithm $ga1$ uses problem-specific search operators as defined in Section \ref{sec:ga_design}. Algorithm $ga2$ uses simulated binary crossover as well as polynomial mutation~\cite{deb2007self}, as defined in Section~\ref{sec:latent_space}, which are state-of-the-art vector-based search operators. They can be used out-of-the box and do not need to be customized for a given problem.
The original space $or$ consists of solutions represented as vectors in $N$-dimensional space, as defined in Section  \ref{sec:ga_design}. 

The second search space is the latent space  ($vr$) obtained by training a VAE on a dataset of 10,000 valid, randomly generated test scenarios. The corresponding search algorithms in this space are referred to as $vr\_ran$, $vr\_ga1$, and $vr\_ga2$.
 Lastly, we consider the latent space of a VAE trained on test scenarios optimized with a simplified function $F_s$ ($vo$), as described in Section~\ref{sec:surrogate_function}. The search algorithms operating in this space are denoted as $vo\_ran$, $vo\_ga1$, and $vo\_ga2$.
For each configuration, we conduct 10 runs in the simulated environments corresponding to the two considered use cases, where the objective functions are evaluated based on the SUT's behavior in simulation. The test generation budget is the same as in RQ1, i.e., 200 evaluations and 2 hours for UC1 and UC2, respectively.

\begin{figure}[h!]
\includegraphics[scale=0.45]{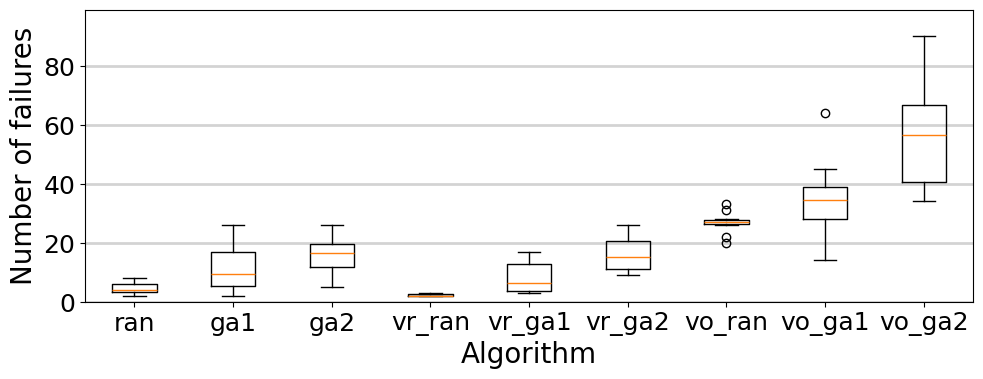}
\centering
\caption{Number of failures for different search algorithms in different search spaces for UAV case study}
\label{fig:rq2-uav-failures}
\end{figure}

\begin{figure}[h!]
\includegraphics[scale=0.45]{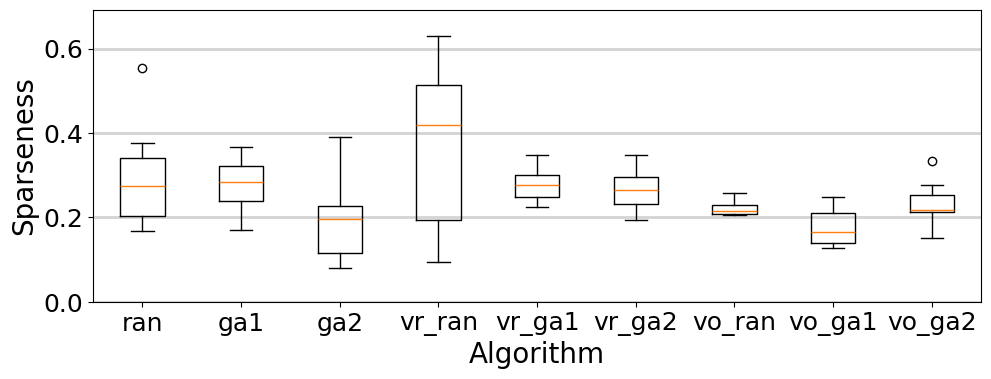}
\centering
\caption{Sparseness of failures for different search algorithms in different search spaces for UAV case study}
\label{fig:rq2-uav-sparseness}
\end{figure}

\begin{figure}[h!]
\includegraphics[scale=0.45]{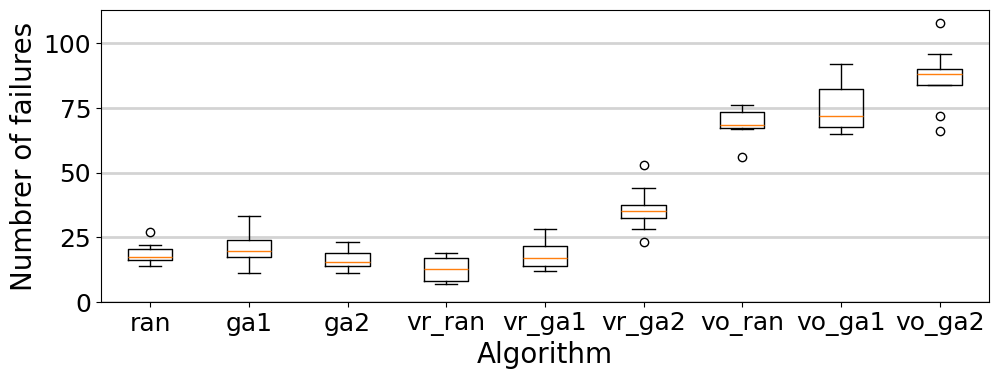}
\centering
\caption{Number of failures for different search algorithms in different search spaces for ADS case study}
\label{fig:rq2-ads-failures}
\end{figure}

\begin{figure}[h!]
\includegraphics[scale=0.45]{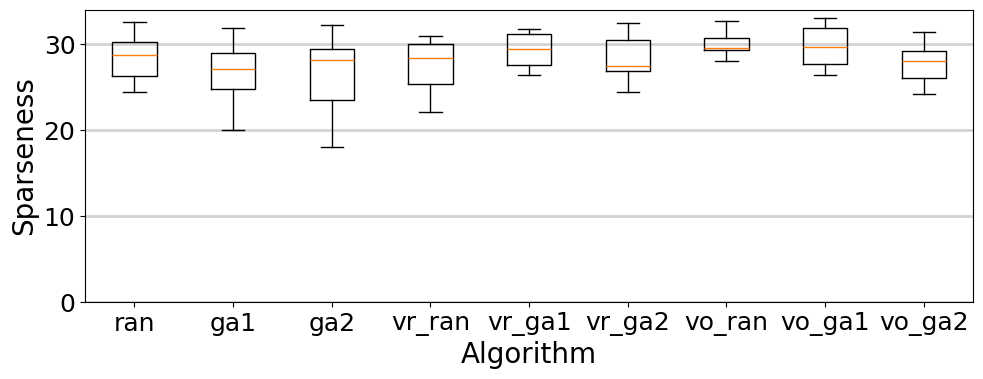}
\centering
\caption{Sparseness of failures for different search algorithms in different search spaces for ADS case study}
\label{fig:rq2-ads-sparseness}
\end{figure}

\textbf{Results.}
Figures~\ref{fig:rq2-uav-failures} and~\ref{fig:rq2-ads-failures} illustrate the number of failures identified by different algorithms for the UAV and ADS case studies, respectively. Similarly, Figures~\ref{fig:rq2-uav-sparseness} and~\ref{fig:rq2-ads-sparseness} present the sparseness of the uncovered failures. Statistical test results are provided in Appendix~\ref{sec:appendix} and summarized in Tables~\ref{tab:rq2-uav} and~\ref{tab:rq2-ads}.

Firstly, we can observe that the search algorithms in the optimized latent space reveal more failures with a large effect size, compared to the search in original and random latent spaces for both use cases. Specifically, in UC1, the best-performing search algorithm in the $vo$ space, $vo\_ga2$, uncovers 3.6 and 3.48 times more failures than the best-performing algorithms in the $or$ and $vr$ spaces, respectively. In UC2, the $vo$ space reveals 2.4 times more failures than $vr$ and 4.14 times more than $or$.

At the same time, the algorithms in all the three spaces produce failures with high diversity, comparable to random generation ($ran$). In the UAV case study, only the $vo\_ga1$ algorithm demonstrates a statistically significant reduction in sparseness, being 66\% lower than $ran$. In the ADS case study, no statistically significant difference in sparseness is observed for algorithms in $vo$ and $vr$ compared to random search. 

In the UAV case study, no clear advantage in terms of revealed failures is observed when comparing the original and random latent spaces. The algorithm $vr\_ga2$ reveals 33\% more failures than $ga1$, with no statistically significant difference in performance compared to $ga2$. However, in the ADS case study, $vr\_ga2$ outperforms the best-performing algorithm in the original space, $ga1$, uncovering 71\% more failures with a large effect size. Notably, $ga1$ relies on problem-specific search operators, which are time-consuming to design and fine-tune. In contrast, $vr\_ga2$ leverages readily available, state-of-the-art search operators used in evolutionary search. This indicates that optimization in random latent space can be a promising approach, achieving effective search performance with standard search operators and without the need for developing problem-specific ones.

The effectiveness of state-of-the-art search operators, such as simulated binary crossover and polynomial mutation, is further validated by the results in random and optimized latent search spaces. In both case studies, a GA employing these search operators outperforms a GA using domain-specific search operators. For the UAV testing, the algorithm $vr\_ga2$ identifies 2.2 times more failures than $vr\_ga1$, while $vo\_ga2$ uncovers 1.56 times more failures than $vo\_ga1$. Similarly, in the ADS use case, $vr\_ga2$ identifies 1.44 times more failures than $vr\_ga1$, and $vo\_ga2$ uncovers 1.15 times more failures than $vo\_ga1$.
Interestingly, in the original space, the domain-specific search operators can be more effective. In the UAV case study, no statistically significant difference is observed between $ga1$ and $ga2$ performance. In the ADS case study, on the other hand, $ga1$ with domain-specific search operators outperforms $ga2$ with medium effect size.

\begin{tcolorbox}
\textbf{Summary of RQ2:} For UC1, searching in the optimized latent space is 3.5 times more effective in revealing failures compared to the original and random latent spaces. For UC2, it is 2.4 times more effective than the random latent space and 4.14 times more effective than the original latent space. Algorithms in all three spaces generate failures with high diversity, comparable to random generation. In the ADS use case, optimization in the random latent space outperforms the original space by 71\%, while no significant difference in the number of failures is observed for the UAV use case. GA with vector-based search operators uncovers more failures in the latent space with a large effect size compared to GA with problem-specific operators. However, in the original space, problem-specific search operators outperform vector-based ones with a medium effect size in the ADS use case.
\end{tcolorbox}
 \subsubsection{RQ3 – Choice of VAE hyperparameters}
\hfill 

 \textbf{Motivation.} Hyperparameter selection is an important step in the development of machine learning models. In this RQ, we aim to understand what VAE architecture and hyperparameters allow us to obtain a VAE with the lowest validation loss. Lower loss indicates that the VAE learned to perform the reconstruction with higher fidelity, which is important during the optimization process \cite{bentley2022coil}. 
 We focus our evaluation mainly on VAE architecture selection in terms of number of hidden layers and their size, which was not previously explored in the literature on optimization in latent space. We also experiment with batch size and learning rate, which are known to be sensitive hyperparameters~\cite{smith2017don}.
\begin{figure}[h!]
\begin{subfigure}{0.45\textwidth}
  \centering
\includegraphics[scale=0.6]{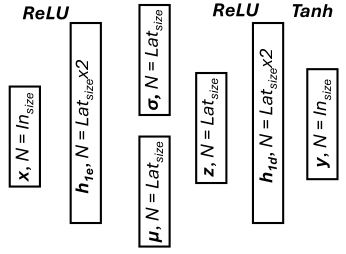} 
  \caption{VAE1 architecture}
  \label{fig:rq3-vae1}
\end{subfigure}
\begin{subfigure}{0.45\textwidth}
  \centering
\includegraphics[scale=0.6]{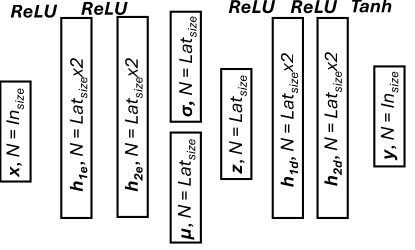}  
  \caption{VAE2 architecture}
  \label{fig:rq3-vae2}
\end{subfigure}
    \vspace{1em} 
    \begin{subfigure}{0.45\textwidth}
    \centering
     \includegraphics[scale=0.6]{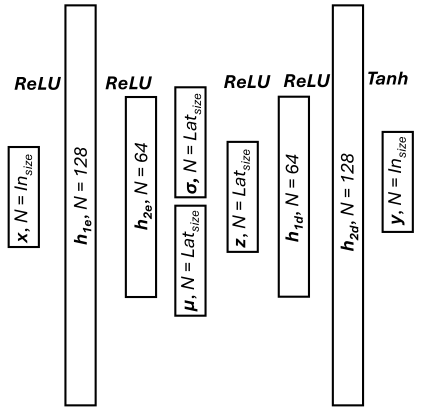}
     \caption{VAE3 architecture}
        \label{fig:rq3-vae3}
    \end{subfigure}
\caption{Different VAE configurations considered in hyperparameter fine-tuning }
\label{fig:rq3-vae}
\end{figure}

 \begin{figure}[h!]
\begin{subfigure}{0.45\textwidth}
  \centering
\includegraphics[scale=0.08]{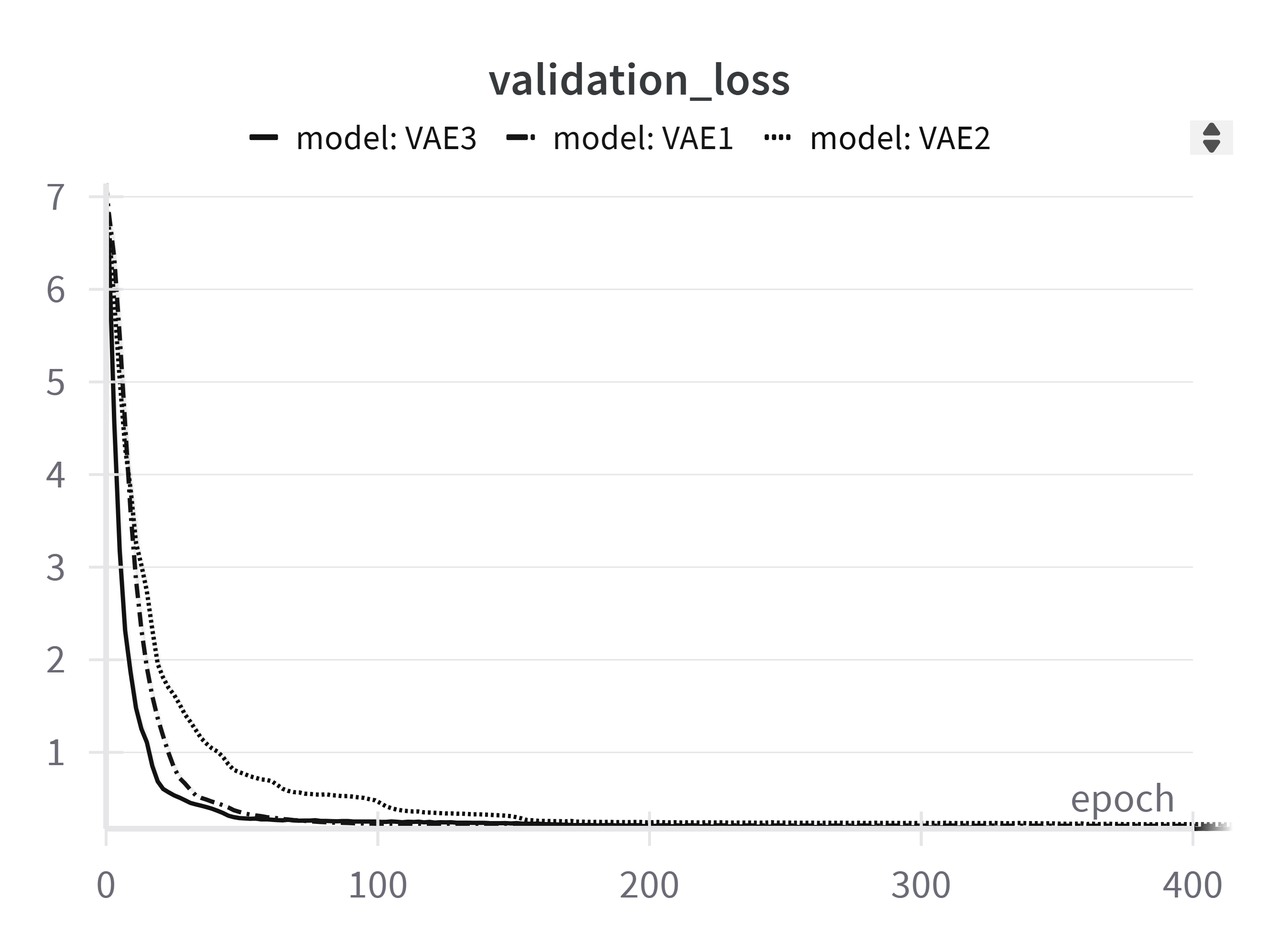} 
  \caption{Convergence of VAE loss in the UAV use case}
  \label{fig:rq3_vae_uav}
\end{subfigure}
\begin{subfigure}{0.45\textwidth}
  \centering
\includegraphics[scale=0.075]{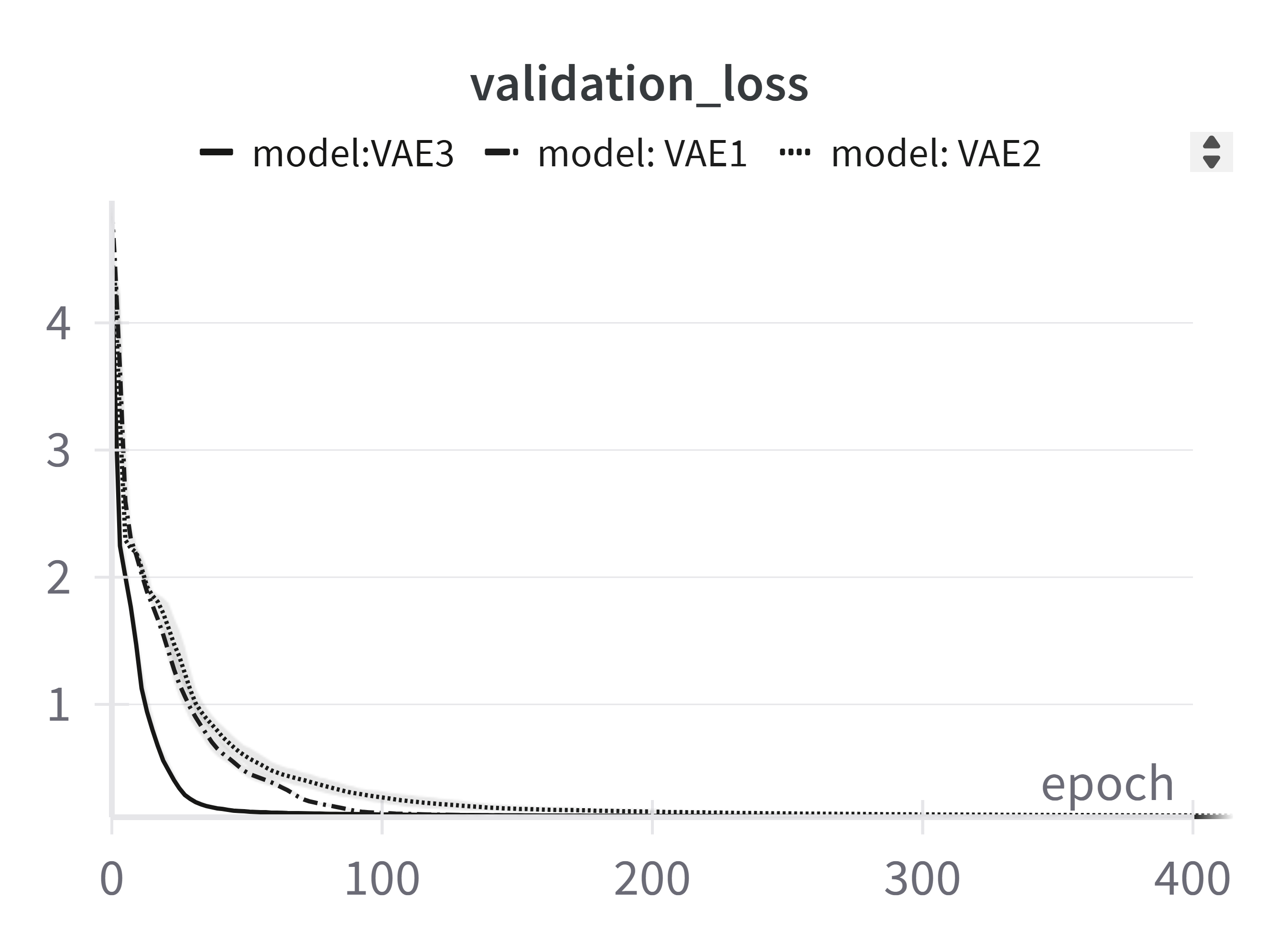}  
  \caption{Convergence of VAE loss in the ADS use case}
  \label{fig:rq3_vae_ads}
\end{subfigure}
\caption{Convergence of VAE loss for different configurations with batch size of 512 and learning rate of 0.001}
\label{fig:vae_train_rq3}
\end{figure}
\begin{table}[h!]
    \centering
    \caption{Hyperparameter selection for VAE in UAV and ADS case studies.}
    \scalebox{0.9}{
    \begin{tabular}{ccccc|ccccc}
    \hline
    \multicolumn{5}{c|}{\textbf{UAV Use Case}} & \multicolumn{5}{c}{\textbf{ADS Use Case}} \\ \hline
    \textbf{Model} & \textbf{LR} & \textbf{Batch Size} & \textbf{Time (s)} & \textbf{Val. Loss} & 
    \textbf{Model} & \textbf{LR} & \textbf{Batch Size} & \textbf{Time (s)} & \textbf{Val. Loss} \\ \hline
    VAE3  & 0.001  & 512  & 1002.87  & \textbf{0.139}  & VAE3  & 0.001  & 512  & 932.17  & \textbf{0.104} \\
    VAE3  & 0.0001 & 512  & 1004.97  & 0.167  & VAE1  & 0.001  & 512  & 883.38  & 0.109 \\
    VAE1  & 0.001  & 512  & 956.34  & 0.194  & VAE2  & 0.001  & 512  & 907.16  & 0.111 \\
    VAE2  & 0.001  & 512  & 975.89  & 0.196  & VAE3  & 0.0001 & 512  & 929.47  & 0.113 \\
    VAE1  & 0.0001 & 512  & 951.70  & 0.203  & VAE1  & 0.0001 & 512  & 882.93  & 0.116 \\
    VAE2  & 0.0001 & 512  & 976.16  & 0.208  & VAE2  & 0.0001 & 512  & 905.43  & 0.122 \\ \hline
    VAE3  & 0.001  & 128  & 1243.11  & 0.388  & VAE3  & 0.001  & 128  & 1170.02  & 0.278 \\
    VAE3  & 0.0001 & 128  & 1245.90  & 0.429  & VAE3  & 0.0001 & 128  & 1157.84  & 0.289 \\
    VAE1  & 0.001  & 128  & 1126.21  & 0.457  & VAE2  & 0.001  & 128  & 1095.28  & 0.297 \\
    VAE2  & 0.001  & 128  & 1197.99  & 0.458  & VAE1  & 0.001  & 128  & 1029.93  & 0.304 \\
    VAE1  & 0.0001 & 128  & 1132.14  & 0.471  & VAE2  & 0.0001 & 128  & 1098.26  & 0.310 \\
    VAE2  & 0.0001 & 128  & 1192.60  & 0.488  & VAE1  & 0.0001 & 128  & 1026.82  & 0.315 \\ \hline
    VAE3  & 0.001  & 64   & 1596.64  & 0.647  & VAE3  & 0.001  & 64   & 1479.13  & 0.451 \\
    VAE3  & 0.0001 & 64   & 1553.38  & 0.681  & VAE3  & 0.0001 & 64   & 1426.66  & 0.465 \\
    VAE2  & 0.001  & 64   & 1422.88  & 0.717  & VAE2  & 0.001  & 64   & 1308.59  & 0.474 \\
    VAE1  & 0.001  & 64   & 1310.35  & 0.738  & VAE1  & 0.001  & 64   & 1207.79  & 0.499 \\
    VAE1  & 0.0001 & 64   & 1310.20  & 0.744  & VAE2  & 0.0001 & 64   & 1301.36  & 0.506 \\
    VAE2  & 0.0001 & 64   & 1428.39  & 0.748  & VAE1  & 0.0001 & 64   & 1200.66  & 0.519 \\ \hline
    \end{tabular}}
    \label{tab:rq3_ads_uav}
\end{table}

 \textbf{Method.} In this RQ, we train VAEs using a dataset of optimized test scenarios. For training, we used Adam optimizer with first and second momentum coefficients set to 0.9 and 0.999, respectively, which are the recommended parameters in the literature \cite{kingma2014adam}. We evaluated VAEs trained with different batch sizes (64, 128, and 512) and learning rates (0.001 and 0.0001). We train the VAE on the optimized dataset of 8000 training and 2000 validation entries.
We considered three VAE architectures with different number of hidden layers and size sizes, as shown in Figure~\ref{fig:rq3-vae}. 
The first VAE configuration, VAE1, is based on the setup proposed by Bentley et al. \cite{bentley2022coil}. It includes one hidden layer and a latent space size that is double the input space size of 17, resulting in approximately 3019 parameters. The second configuration, VAE2, adds an additional hidden layer, nearly doubling the number of parameters to 5889.
The third configuration, VAE3, consists of two hidden layers with sizes 128 and 64, respectively, and has a total of approximately four times the parameters of VAE2, i.e., 22435 parameters.
Each hidden layer is followed by a Relu activation function. The inputs are normalized to the range between [-1, 1], and the $Tanh$ function is used at the output layer.
In total, we evaluated 18 VAE configurations. For each configuration, VAE was trained three times for 1000 epochs. We report the average metrics, such as validation loss observed at the final epoch and training time.
In all experiments, the size of the latent space \textbf{$z$} was fixed and equal to the input size.

 \textbf{Results.} The obtained results are presented in Table~\ref{tab:rq3_ads_uav}. We observe that in both case studies, the configuration with $VAE3$, a learning rate of 0.001, and a batch size of 512 results in the lowest validation loss. Batch size has an important effect on the final loss and training time. For instance, in the UAV use case, with batch size decreasing from 512 to 128 and the rest of the parameters remaining fixed, the validation loss increases 2.79 times, and training time increases by 25\%. Similarly, in the ADS use case, the validation loss increases 2.67 times and the training time increases by 25.5\%. The increase in training time for lower batch sizes can be attributed to the additional computations required to calculate gradients and update the weights more frequently.

Examining the differences in VAE configurations for the UAV use case, the lowest loss achieved by VAE3 is 28.3\% lower than that of VAE1 and 29\% lower than that of VAE2. Similarly, in the ADS use case, VAE3 achieves a 4.5\% lower loss than VAE1 and 6.4\% lower loss than VAE2. 
Furthermore, as shown in Figure~\ref{fig:vae_train_rq3}, where all three VAE configurations use the same learning rate (0.001) and batch size (512), VAE3 demonstrates the fastest convergence.

\begin{tcolorbox}
\textbf{Summary of RQ3:} Overall, the best performance was observed with larger batch sizes (512) and learning rates of 0.001 for both case studies, with VAE configuration having bigger layer dimensions showing faster convergence and smaller final loss.

\end{tcolorbox}

 \subsubsection{RQ4 – Impact of the number of latent variables on the VAE performance}
\hfill

\textbf{Motivation.}
One of the key components of a VAE is the latent space. Maintaining the latent space dimension equal to the input size increases the likelihood of achieving a lower reconstruction error. However, reducing the latent space dimension decreases the problem's complexity, making optimization easier~\cite{bentley2023using}. In this research question, we aim to explore the trade-off between latent space dimensionality and reconstruction fidelity.

\textbf{Method.} Having selected the key hyperparameters in RQ3, i.e., VAE3 configuration, 512 batch size, and 0.001 learning rate, we train VAEs with 10 different latent space sizes $nlat$, ranging from 8 to 25, for 1000 epochs. In each configuration, we train the VAE 3 times and select the VAE with the lowest validation loss ($l_v$). 
To verify that the VAE has learned a meaningful representation, we compare the distances between the initial test scenarios in the validation set (2,000 scenarios) and their reconstructed versions, i.e., the respective inputs and outputs of the VAE. Unlike validation loss, which considers both reconstruction error and KLD loss, distance metric gives a better evaluation of the fidelity of reconstruction. We use cosine distance $c_d$ that we previously applied for duplicate removal and sparseness evaluation in the UAV case study. Experimentally, we established a threshold of 0.025 for identifying scenarios as duplicates. 
We thus expect the cosine distance between the input and reconstructed VAE test scenarios to be less than the established threshold. Otherwise, it signifies that the VAE has not learned a sufficiently meaningful representation and that reconstructed test cases differ from the original ones.
We collect these metrics for two VAE types: one trained with a randomized dataset ($VAE_{R}$) and the other with the optimized dataset ($VAE_{O}$).

\begin{figure}[h!]
\begin{subfigure}{0.45\textwidth}
  \centering
\includegraphics[scale=0.45]{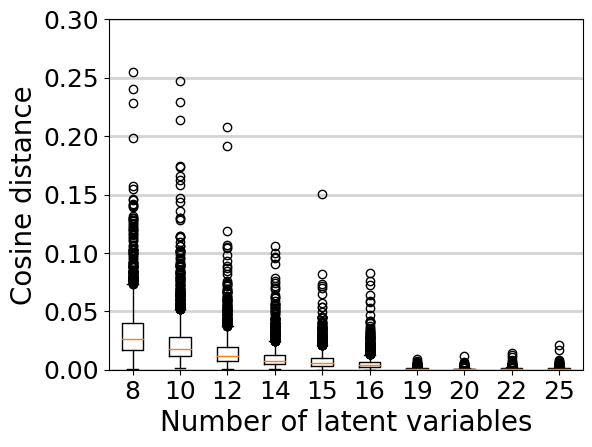} 
  \caption{Cosine distance for the UAV case study}
  \label{fig:rq4_uav_r}
\end{subfigure}
\begin{subfigure}{0.45\textwidth}
  \centering
  \includegraphics[scale=0.45]{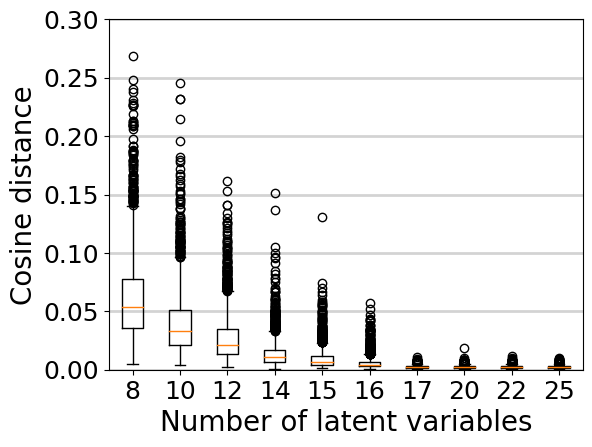}  
  \caption{Cosine distance for the ADS case study}
  \label{fig:rq4_ads_r}
\end{subfigure}
\caption{Cosine distance between original tests and their reconstruction by the VAE based on a random dataset}
\label{fig:rq4_r}
\end{figure}

\textbf{Results.} The boxplots of cosine distance between original and reconstructed tests in validation set for $VAE_R$ are shown in Figure \ref{fig:rq4_r}. The values for the average cosine distance, validation loss, and training time are shown in Table~\ref{tab:rq4_random}. We can see that for both use cases, the average cosine distance and validation loss decrease with bigger latent space size. At the same time, when the latent space dimension reaches the same size as the input, the validation loss and cosine distance values stabilize and don't show a noticeable reduction. The training time remains relatively stable, with a slight increase for latent space sizes, which are bigger than the initial ones. 
The boxplots further show that while the average cosine distance generally remains below the defined threshold, some reconstructions have cosine distances exceeding the threshold for latent dimensions of 16 or smaller. This suggests that $VAE_R$ struggles to learn meaningful representations for certain tests when the latent space has a lower dimensionality.

\begin{table}[!ht]
    \centering
    \caption{Evaluation of different latent space dimensions for random dataset based VAE}
    \scalebox{0.85}{
    \begin{tabular}{ccccc|cccc}
    \hline
        \textbf{UAV use case} & ~ & ~ & ~ & ~ & \textbf{ADS use case} & ~ & ~ & ~ \\ \hline
        \textbf{nlat} & \textbf{av dist} & \textbf{max dist} & \textbf{validation loss} & \textbf{training time} & \textbf{av dist} & \textbf{max dist} & \textbf{validation loss} & \textbf{training time} \\ 
        8 & 0.03205 & 0.2551 & 1.7417 & 530.68 & 0.06236 & 0.333 & 0.6998 & 526.87 \\ 
        10 & 0.02362 & 0.2468 & 1.3662 & 530.02 & 0.04059 & 0.2454 & 0.5042 & 528.5 \\ 
        12 & 0.0159 & 0.2075 & 0.9356 & 528.51 & 0.02726 & 0.1619 & 0.3547 & 527.41 \\ 
        14 & 0.01077 & 0.1061 & 0.6649 & 531.06 & 0.01422 & 0.1515 & 0.2315 & 528.36 \\ 
        15 & 0.00838 & 0.1509 & 0.526 & 528.86 & 0.00967 & 0.1307 & 0.182 & 527.4 \\ 
        16 & 0.00568 & 0.0831 & 0.3782 &\textbf{ 527.15} & 0.00608 & 0.0568 & 0.1478 & 526.79 \\ 
        19 & 0.0009 & \textbf{0.0088} & 0.1604 & 527.35 & \textbf{0.00249 }& 0.0113 & \textbf{0.1112 }& 526.86 \\ 
        20 & \textbf{0.00085} & 0.0115 & \textbf{0.1596} & 532.94 & 0.00255 & 0.019 & 0.1116 & \textbf{524.98} \\ 
        22 & 0.00099 & 0.0141 & 0.1614 & 550.4 & 0.0026 & 0.0114 & 0.1117 & 536.55 \\ 
        25 & 0.001 & 0.0214 & 0.1667 & 539.43 & 0.00257 & \textbf{0.0099 }& 0.1116 & 531.21 \\ \hline
    \end{tabular}}
    \label{tab:rq4_random}
\end{table}

\begin{figure}[h!]
\begin{subfigure}{0.45\textwidth}
  \centering
\includegraphics[scale=0.45]{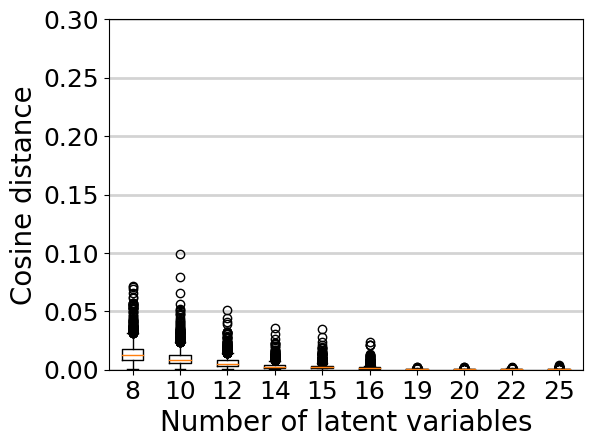} 
  \caption{Cosine distance for the UAV case study}
  \label{fig:rq4_uav_o}
\end{subfigure}
\begin{subfigure}{0.45\textwidth}
  \centering
  \includegraphics[scale=0.45]{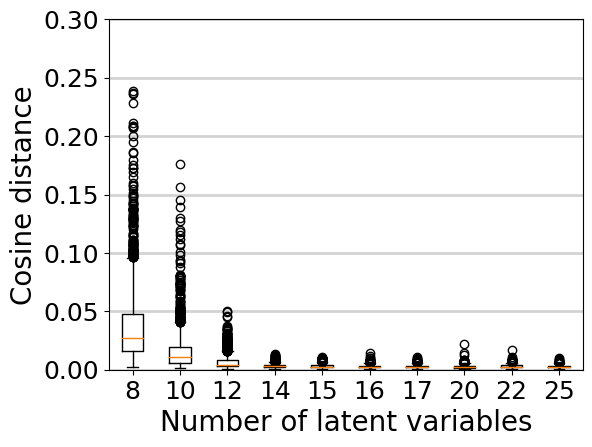}  
  \caption{Cosine distance for the ADS case study}
  \label{fig:rq4_ads_o}
\end{subfigure}
\caption{Cosine distance for VAE based on optimized dataset with different sizes of the latent space}
\label{fig:rq4_o}
\end{figure}
\begin{table}[!ht]
    \centering
    \caption{Evaluation of different latent space dimensions for optimized dataset based VAE}
    \scalebox{0.85}{
    \begin{tabular}{ccccc|cccc}
    \hline
        \textbf{UAV use case} & ~ & ~ & ~ & ~ & \textbf{ADS use case} & ~ & ~ & ~ \\ \hline
        \textbf{nlat} & \textbf{av dist} & \textbf{max dist} & \textbf{validation loss} & \textbf{training time} & \textbf{av dist} & \textbf{max dist} & \textbf{validation loss} & \textbf{training time} \\ 
        8 & 0.01443 & 0.072 & 1.219 & 532.048 & 0.0375 & 0.312 & 0.4686 & 529.469 \\ 
        10 & 0.01045 & 0.0994 & 0.866 & 531.720 & 0.0157 & 0.176 & 0.241 & 528.835 \\ 
        12 & 0.00642 & 0.0513 & 0.548 & 529.740 & 0.0065 & 0.05 & 0.1343 & 529.694 \\ 
        14 & 0.00347 & 0.036 & 0.363 & 531.321 & 0.0032 & 0.013 & 0.1103 & 530.276 \\ 
        15 & 0.00264 & 0.035 & 0.289 & 528.464 & 0.003 & 0.011 & \textbf{0.1072} & 528.498 \\ 
        16 & 0.00177 & 0.0241 & 0.22 & \textbf{527.025} & 0.0029 & 0.014 & 0.108 & \textbf{519.634} \\ 
        19 & 0.00059 & 0.0023 & 0.138 & 527.828 & 0.0029 & 0.011 & 0.1082 & 520.837 \\ 
        20 &\textbf{ 0.00049} & 0.0023 & 0.138 & 553.149 & \textbf{0.0027} & 0.022 & 0.1077 & 551.710 \\ 
        22 & 0.00061 & \textbf{0.002} & 0.141 & 563.320 & 0.0029 & 0.017 & 0.1079 & 534.286 \\ 
        25 & 0.00051 & 0.0044 &\textbf{ 0.136} & 549.533 & 0.0029 & \textbf{0.01} & 0.1086 & 525.334 \\ \hline
    \end{tabular}}
    \label{tab:rq4_optimized}
\end{table}

From Figure \ref{fig:rq4_o} and Table \ref{tab:rq4_optimized}, which present the results for the VAE trained with the optimized dataset ($VAE_O$), we observe a similar trend to that of $VAE_R$. Specifically, the validation loss and cosine distance decrease as the number of latent variables increases, eventually plateauing when the number of latent variables ($nlat$) approaches the input size. However, for the same value of $nlat$, the average validation loss and cosine distance are lower for $VAE_O$. For instance, given $nlat=19$, in the UAV use case $l_v$ is 14 \% lower for $VAE_O$ than $VAE_R$ and $d_c$ is 34.4 \% lower. In the ADS use case, $v_l$ is 2.7 \% lower and no significant difference is observed for $d_c$. Bigger differences in $l_v$ and $d_c$ compared to respective values in $VAE_R$ are observed from smaller values of $nlat$. 
In particular, for the UAV use case, the VAE with up to 16 latent dimensions (15.7 \% reduction in size) does not exceed the $c_d$ threshold. Similarly, in the ADS use case, the threshold is not exceeded for up to 14 (17.6 \% reduction) latent dimensions. This indicates that a big portion of tests can be accurately represented with a smaller number of variables, simplifying the optimization process.
Intuitively, the dataset for $VAE_O$ contains less variety of inputs than $VAE_R$, making it easier to learn the representation.
Finally, increasing the number of latent dimensions more than the original input size does not have a noticeable effect on reducing average $v_l$ or $c_d$.

\begin{tcolorbox}
\textbf{Summary of RQ4:} 
Reducing the number of latent variables in a VAE increases both the validation loss and the cosine distance between the original and reconstructed tests. Conversely, increasing the number of latent variables beyond the input size does not significantly improve reconstruction accuracy. However, when an optimized dataset is used, the latent space size can be reduced by 15.7\% for UAV testing and 17.6\% for ADS without any noticeable reduction in reconstruction quality.  
\end{tcolorbox}
\subsubsection{RQ5 – RILaST overhead}
\hfill

\textbf{Motivation.} In this RQ we evaluate the additional time overhead related to our RILaST approach. This includes the time to collect the dataset, train the VAE and perform the inference.

\textbf{Method.} For each use case, we record the time needed to generate random and optimized datasets, VAE training, and inference time. To measure the random dataset creation time, we generated 10 randomized datasets and recorded the average generation time.
For the optimized dataset, to collect a dataset of 10,000 entries, we ran the genetic algorithm 50 times with a population of 200 for 50 generations, as described in Section~\ref{sec:RILaST}. We recorded the average time it took to complete each run.

To evaluate VAE training time, we trained VAEs for each use case based on randomized and optimized datasets with latent space size corresponding to the input size. We ran the training 5 times for each configuration and recorded the average training time values.
To record inference time, we used the trained VAE in each configuration during the optimization process and measured the time it takes to decode a test. We recorded the values for 100 inferences. In the results section, we report the average value of time taken as well as the standard deviation (in brackets). 

\begin{table}[!ht]
    \centering
    \caption{Overhead of training VAE with random generated dataset}
    \scalebox{0.9}{
    \begin{tabular}{ccc|ccc}
    \hline
        UAV use case & ~ & ~ & ADS use case & ~ & ~ \\ \hline
        Dataset collection, s  & VAE training, s & VAE inference, s & Dataset collection, s  & VAE training, s & VAE inference, s \\ 
        41.692 (1.85) & 527.34 (3.17) & 0.0011 (0.00021) & 90.32 (2.13) & 526.86 (2.796) & 0.00107 (0.00023) \\ \hline
    \end{tabular}}
    \label{tab:rq5_random}
\end{table}
\begin{table}[!ht]
    \centering
    \caption{Overhead of training VAE with optimized  dataset}
    \scalebox{0.9}{    \begin{tabular}{ccc|ccc}
    \hline
        UAV use case & ~ & ~ & ADS use case & ~ & ~ \\ \hline
        Dataset collection, x0.02 s  & VAE training, s & VAE inference , s & Dataset collection, x0.02 s  & VAE training, s & VAE inference , s \\ 
        338.52 (53.173) & 527.828 (2.16) & 0.0011(0.00026) & 66.732 (7.008) & 520.83 (10.27) & 0.00115 (0.00038) \\ \hline
    \end{tabular}}
    \label{tab:rq5_optimized}
\end{table}
\textbf{Results}. From Table~\ref{tab:rq5_random} we can see that the biggest time overhead when training the VAE with randomized dataset is in the dataset collection phase, taking on average 527 seconds in both use cases. The data collection time is around 42 seconds for the UAV use case, taking more than twice as long (90 seconds) in the ADS use case. Data inference time takes on average 0.001 s per evaluation in both use cases. Considering that typically no more than 1000 evaluations will be used in a single run, this time remains negligible.

For VAE with optimized dataset, the biggest overhead is in the dataset collection phase, as it can be seen from Table~\ref{tab:rq5_optimized}. Genetic algorithm with 50 generations and a population size of 200 takes an average of 338.5 seconds per run in the UAV use case. Given a dataset size of 10,000, fifty runs are required, resulting in a total collection time of approximately 4.7 hours. In contrast, data collection for the ADS use case is five times faster, with each run taking an average of 66 seconds and a total dataset collection time of about 0.9 hours. We attribute this to the fact that the simplified objective function for ADS is less computationally expensive.
Although the dataset collection time is substantial, it can be significantly reduced, as genetic algorithm runs are highly parallelizable. Currently, no parallelization techniques have been applied. The VAE training times are comparable to those observed with the random dataset-based VAE, averaging approximately 527 seconds. VAE inference remains minimal, requiring only 0.001 seconds per evaluation.

\begin{tcolorbox}
\textbf{Summary of RQ5:} 
VAE training introduces the largest time overhead, with the random dataset-based VAE requiring 527 seconds on average. Dataset collection times are relatively low, ranging from 40 to 90 seconds. Inference time is negligible, taking less than 10 milliseconds. However, for the optimized dataset-based VAE, dataset collection time can be substantial, depending on the computational cost of the simplified objective function. In the UAV use case, collection took 4.7 hours, whereas in the ADS case study, it required only 0.9 hours. This time can be significantly reduced, as the genetic algorithms used for dataset collection are highly parallelizable.
\end{tcolorbox}
\subsection{Threats to validity}

\textbf{Internal validity.} 
To minimize the threats to internal validity, relating to experimental errors and biases, whenever available, we used standardized frameworks for development and evaluation. We implemented the evolutionary search algorithms based on a popular Python-based Pymoo framework~\cite{pymoo}. 
We trained the VAE based on a commonly used machine learning framework Pytorch~\cite{paszke2019pytorch}. For the autonomous drone use case, we used  Aerialist benchmark proposed by Khatiri et al.~\cite{icse2024Aerialist}, which was also utilized at the SBFT 2024 UAV tool competition~\cite{khatiri2024sbft}. To evaluate the scenarios for the LKAS case study, we leveraged a standardized test pipeline from the SBST 2022 workshop tool competition~\cite{gambi2022sbst}. 
To minimize the threats to internal validity stemming from selected parameter values, we conducted extensive experimentation with various hyperparameters and configurations for VAE training.  
 
\textbf{Conclusion validity.}
 Conclusion validity is related to random variations and inappropriate
use of statistics. To mitigate it, we followed the guidelines in~\cite{arcuri2014hitchhiker} for search-based algorithm evaluation. We repeat all evaluations a sufficient number of times to ensure statistical significance. For all results, we report the non-parametric Mann-Whitney U test with a significance level of $\alpha = 0.05$, as well as the effect size measure in terms of Cliff's delta. 

\textbf{Construct validity.} Construct validity is related to the degree to which an evaluation measures what it claims. In our experiments, we used two established evaluation metrics: number of revealed failures and their sparseness. For ADS, we used the failure and sparseness definition established in SBST 2022 CPS testing tool competition. For the autonomous UAV, we used the same failure metric definition as in SBFT 2024 UAV test generation competition.
For the evaluation of diversity, we utilized cosine distance, which is a known metric for comparing the difference between vectors~\cite{han2022data}. Prior to using the metric, we ensured that the vectors representing failed test scenarios are properly normalized. Finally, we conclude that our proposed RILaST approach outperforms the available baselines in terms of the number of revealed failures. We acknowledge that this finding is true, provided that RILaST is allocated a one-time budget to generate the dataset and train the VAE.

\textbf{External validity.} External validity relates to the generalizability of our results. We demonstrated how our framework can be applied to the generation of test environments for two different autonomous robotic systems and two test generation tasks. However, we only considered a limited number of test subjects and limited levels of environment complexity. To further explore the generalizability of RILaST, additional experiments with a broader range of robotic systems and more complex environments would be beneficial.

\subsection{Data availability}
The replication package of our experiments, including the implementation of RILaST and the baseline algorithms is available at Zenodo repository~\cite{humeniuk_2025_15087028} as well as GitHub: \url{https://github.com/swat-lab-optimization/RILaST}.

\section{Discussion}\label{sec:discussion}
This section presents the key lessons learned and main recommendations for using RILaST. It also outlines the limitations of RILaST and the challenges encountered when training VAEs for test generation.
\subsection{Usefulness of RILaST and main insights}
As seen from the Results section, performing the search in latent space is beneficial for test generation.
The simplest way to apply RILaST is by training a VAE using a randomly generated dataset, which incurs a low overhead of approximately 10 minutes on average. By leveraging the trained VAE, optimization can be performed in the latent space, eliminating the need to design problem-specific search operators. Instead, readily available vector-based search operators can be used directly. RILaST achieves comparable performance to genetic algorithms (GA) with custom search operators, and can outperform them by up to 71\% in terms of the number of failures revealed.
 
When a simplified fitness function is available, it can guide dataset collection for VAE training, biasing the latent space toward higher-performing solutions and further facilitating the search. This approach increases the number of revealed failures from 4.14 to 4.68 times compared to the original space without compromising diversity. While training a VAE on an optimized dataset may introduce a noticeable overhead, it is a one-time cost that can be significantly reduced using parallelization techniques, such as running the GA in different computational threads for dataset collection.
 
When an optimized dataset is used, the dimensionality of the search space can be reduced from 15.7 \% to 17.6 \% facilitating the search process without introducing noticeable reconstruction errors. 
When training a VAE, architectures with bigger hidden layers converge faster to lower loss values, although no significant difference is observed in the final loss. Learning rate of 0.001 and batch size of 512 show the best performance.

\subsection{Interpreting VAE latent dimensions}
\begin{figure}[h!]
\includegraphics[scale=0.55]{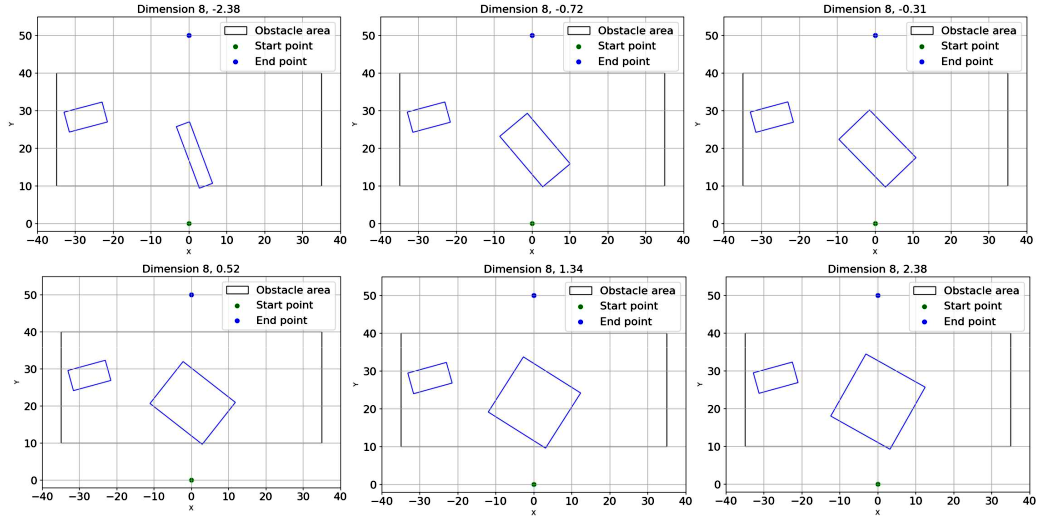}
\centering
\caption{Representation in a latent space for different values of dimension 8 in the UAV use case }
\label{fig:disc-uav}
\end{figure}

The purpose of training the VAE is to learn meaningful representations of tests in the latent space. To visually analyze the learned representation, we sampled random tests in the latent space and fixed the values of all dimensions except one, which we varied within the range of -3$\sigma$ - 3$\sigma$. Figure~\ref{fig:disc-uav} demonstrates an example of the learned representation of a test in the UAV case study. By varying the value of the 8th dimension from -2.38 to 2.38, we observe changes in an obstacle's rotation and width. This suggests that modifying a single latent variable can correspond to transformations involving multiple variables in a latent space.

\begin{figure}[h!]
\includegraphics[scale=0.55]{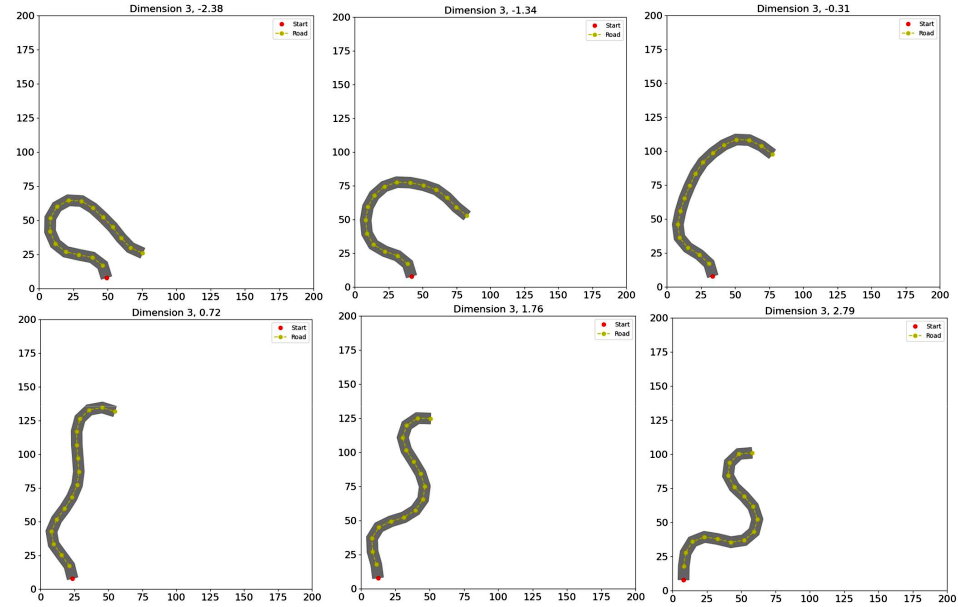}
\centering
\caption{Representation in a latent space for different values of dimension 3 in the ADS use case}
\label{fig:disc-ads}
\end{figure}

Figure \ref{fig:disc-ads} shows an example of the learned representation in UC2, when the dimension 3 is varied. We can observe a complex transformation, where the `left' turn transitions to a `right' turn. Such transformation would involve some complex manipulation of a number of values in the original space (curvature values).
Based on these examples, we argue that the latent space representation can facilitate non-linear transformations of test scenarios that would be difficult to achieve in the original space. This approach could potentially enable more complex transformations, such as modifying non-rectangular obstacle shapes. Additionally, it may be useful for transforming test scenarios with complex representations, such as 3D meshes.

\subsection{Limitations of RILaST}
\begin{figure}[h!]
\begin{subfigure}{0.45\textwidth}
  \centering
\includegraphics[scale=0.35]{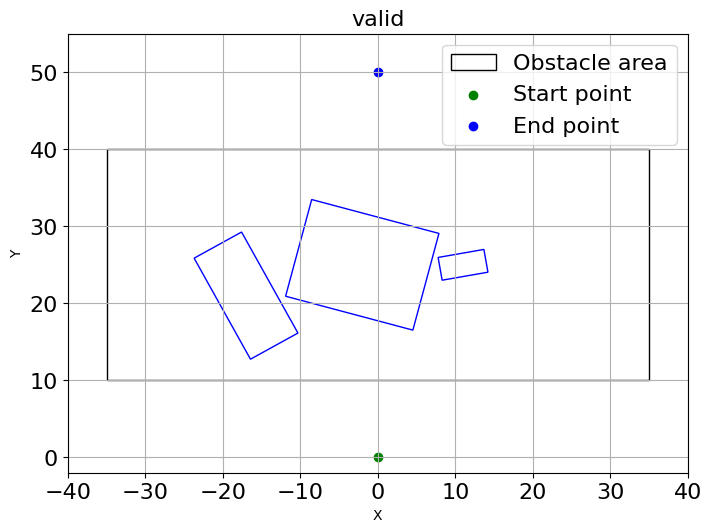} 
  \caption{Original test scenario for UAV (valid)}
  \label{fig:d_uav_v}
\end{subfigure}
\begin{subfigure}{0.45\textwidth}
  \centering
  \includegraphics[scale=0.35]{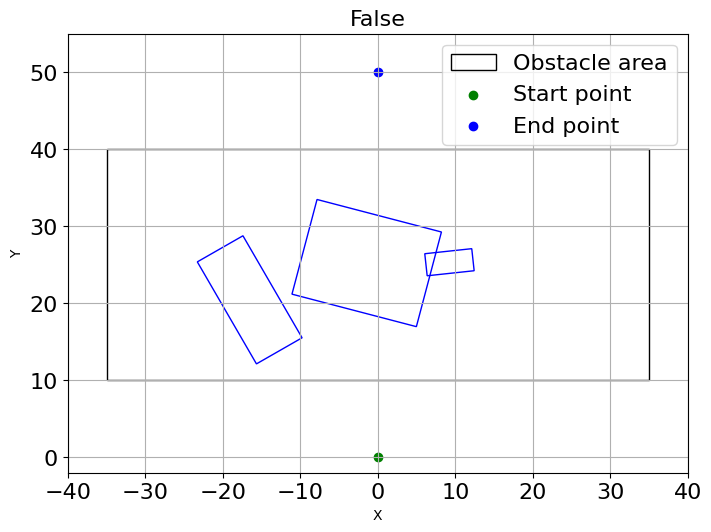}  
  \caption{Reconstructed test scenario for UAV (invalid)}
  \label{fig:d_uav_i}
\end{subfigure}
\caption{Example of a valid original scenario and invalid reconstructed for the UAV use case}
\label{fig:reconstruction_uav}
\end{figure}

\begin{figure}[h!]
\begin{subfigure}{0.45\textwidth}
  \centering
\includegraphics[scale=0.35]{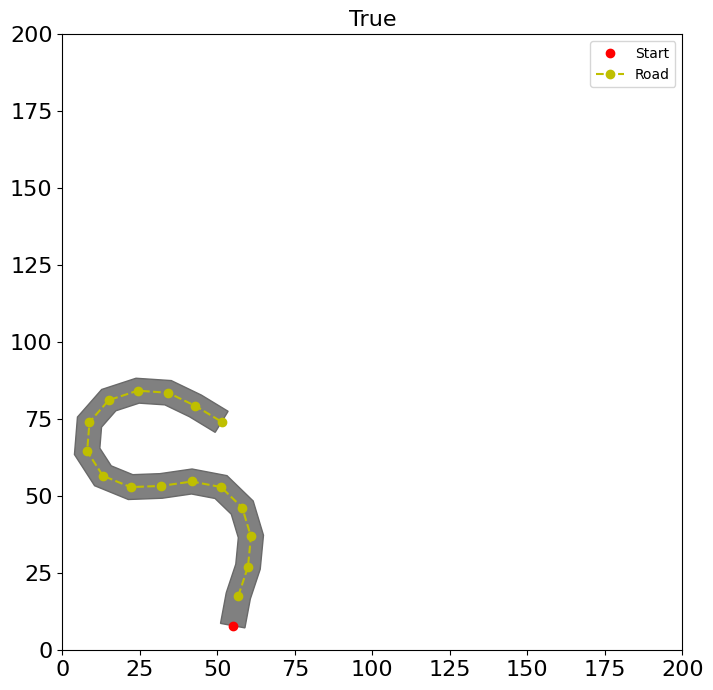} 
  \caption{Original test scenario for ADS (valid)}
  \label{fig:d_ads_v}
\end{subfigure}
\begin{subfigure}{0.45\textwidth}
  \centering
  \includegraphics[scale=0.35]{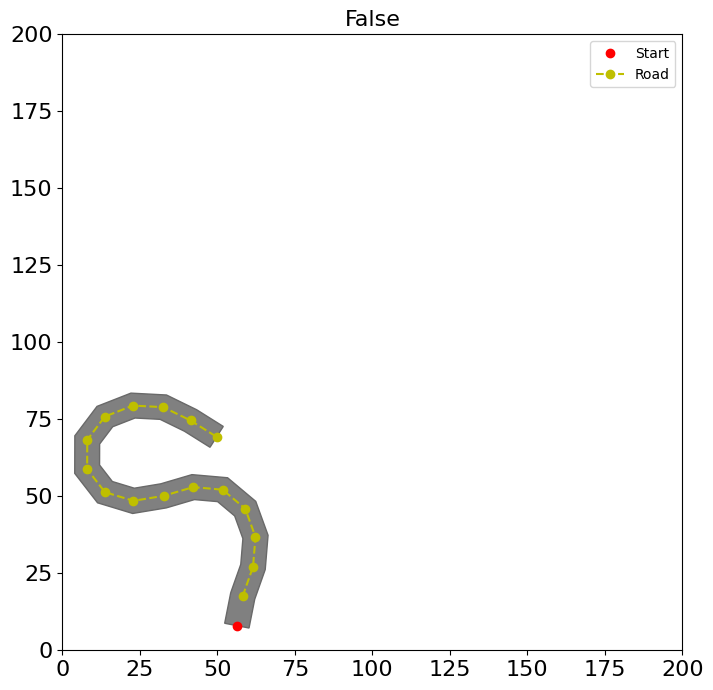}  
  \caption{Reconstructed test scenario for ADS (invalid)}
  \label{fig:d_ads_i}
\end{subfigure}
\caption{Example of a valid original scenario and invalid reconstructed for the ADS use case}
\label{fig:reconstruction_ads}
\end{figure}

In this subsection, we discuss key challenges related to test generation in the latent space. One of the main challenges is designing the loss function for training the VAE. The loss consists of two components: one accounting for reconstruction error and another enforcing Kullback–Leibler divergence to ensure that the latent space follows a normal distribution. However, in test generation for autonomous robotic systems, it is crucial to ensure the validity of the generated tests. In other words, the generated tests must satisfy the pre-defined constraints 
$C$. Current VAE training techniques do not directly account for invalid tests and may result in a latent space containing invalid tests. Figures~\ref{fig:reconstruction_uav} and~\ref{fig:reconstruction_ads} show examples of original (left) and VAE-reconstructed (right) tests that show high similarity and consequently a low reconstruction loss. However, while the original tests on the left are valid, the reconstructed tests on the right are invalid, as they do not satisfy the constraints. In Figure~\ref{fig:d_uav_i} we can see intersecting obstacles and in Figure~\ref{fig:d_ads_i} the produced road topology is too sharp.
We thus argue that there is a need to modify the loss functions by adding regularization terms specific to constraint satisfaction in order to encourage the VAE to learn only valid tests. 
Invalid tests produced in a latent space are not critical, as oftentimes there is an inexpensive function available that checks for test validity prior to its execution in the simulator. We surmise that having a latent space of only valid test scenarios could further improve the search and test generation in a latent space. 

\section{Related work}\label{sec:literature}

In the first part of this section, we discuss the problem of optimization in the latent space in evolutionary search literature. In the second part, we discuss the related works from CPS testing domain.

\subsection{Representation improvement in latent space}
Gaier et al.~\cite{gaier2020discovering} proposed Data-Driven Encoding MAP-Elites (DDE Elites), which is one of the first works to augment the evolutionary search with the search in latent space. The authors describe an approach for generating a dataset of high-performing and at the same time diverse solutions. They leverage MAP-Elites algorithm~\cite{mouret2015illuminating} in combination with the multi-armed bandit strategy to manage the exploration-exploitation trade-off. They show the effectiveness of their approach on the control problem of a 2D planar arm with 20, 200, and 1000 degrees of freedom (Dof). In their experiments, the dimensionality of the latent space is 10, 32, and 200 corresponding to each Dof.

Bentley	et al.~\cite{bentley2022coil} propose Constrained Optimization in Latent Space (COIL) approach, where VAE is used to learn a new representation for search problems that is biased towards solutions that satisfy the problem constraints. Unlike DDE-Elites approach, the learned representations do not involve a reduction of the number of parameters; instead, the main goal of COIL is to remap a hard-to-search genotype space into a more focused easier-to-search latent space. To generate the dataset for VAE training, the authors relied on multiple runs of GA, where constraint satisfaction degree was used as an objective function. The search in the latent space is performed with GA with a simplified configuration, since the latent space is optimized. However, in all the conducted experiments, the authors used only a simple objective function in the form of the squared sum of variables with two different constraints. In their recent work~\cite{bentley2023using} Bentley et al. evaluate COIL on the real world optimization problem of configurations of autonomous robots for last-mile delivery. COIL was able to find valid solutions for all problem variants, outperforming standard GA. The authors further extended their work and proposed the SOLVE: Search space Optimization with Latent Variable Evolution approach \cite{bentley2022evolving}. In SOLVE, the main focus is to bias the representations toward specific criteria or heuristics, including, but not limited to, the satisfaction of constraints. However, SOLVE does not guarantee constraint satisfaction, as its primary goal is search space optimization. SOLVE has only been evaluated using a simple objective function with 5 distinct constraints.
Unlike SOLVE, RILaST focuses not only on optimizing the search space but also on ensuring the generation of diverse and failure-revealing test scenarios in the context of using a computationally expensive simulator-based objective function.

\subsection{Representation improvement for test generation}
In the CPS testing literature, research on representation improvement remains limited.
A number of works focused on the problem of representing road topologies for the purpose of autonomous LKAS system testing. Gambi et al.~\cite{gambi2019automatically} in their AsFault approach propose building a graph representation of the road network, in which edges model road segments and nodes model either road intersections (internal nodes) or intersections between roads and map boundaries. This representation is better suited for evolving road networks than working with polylines. Castellano et al.~\cite{castellano2021analysis} empirically compared six different representations allowing to map road representation in 2D space to a 1D space of values. The results showed that the representation based on a sequence of curvature values of Frenet frames~\cite{tapp2016differential}, referred to as `kappa' representation, outperforms the other alternatives based on Catmull Rom splines and Bezier curves.

Peltomäki et al.~\cite{peltomaki2024learning} propose the WOGAN approach, which utilizes a WGAN~\cite{arjovsky2017wasserstein} architecture to learn the distribution of tests that align with the distribution of safety-critical tests for autonomous systems. However, no search is performed within the learned distribution; tests are only sampled from it and prioritized based on the `analyzer' module predictions.
Winsten et al.~\cite{winsten2024adaptive} apply WOGAN for test generation in autonomous UAVs. To prevent the generation of invalid test scenarios when training the WGAN generative model, the authors impose several complex geometrical constraints on the test representation. This is similar to the optimized dataset collection phase in RILaST, where an heuristic is used to guide the data collection. However, RILaST goes one step further and performs the search in the latent space of the generative model. 

Another group of approaches is focused on the generation of traffic scenarios.
Lu et al.~\cite{lu2024epitester} propose an approach with adaptive representation named EpiTESTER. In EpiTESTER, the test scenario representation is evolving along with the test parameters, making some parameters more important in the search than the others. This approach is based on the epiGA~\cite{epiga} algorithm, which keeps a population of chromosomes and corresponding nucleosomes, that indicate what genes can be mutated. The intuition is that during testing, as the environment changes, some scenario parameters become more important than others. 
Huai et al.~\cite{huai2023sceno} develop the SCENORITA approach, where the main contribution is in introducing the so called ``fully mutable obstacles''. These are obstacles whose parameters can be modified not only during the mutation stage, but also during the crossover. This is achieved by representing obstacles as individuals instead of genes. These approaches focus on representation improvement in the original space, while RILaST operates in a learned latent space, which inherently captures structural relationships between test scenarios.

\section{Conclusions}\label{sec:conclusions}

In this paper, we presented a novel approach, \textbf{RILaST}, for test representation enhancement in a latent space. Based on experiments with two systems under test, a UAV obstacle avoidance system and a lane keeping assist system, we have shown that search in an optimized latent space leads to the discovery of more failures compared to search in the original space.
Moreover, representing tests in a latent space enables the learning of non-linear test transformations, which would otherwise require complex processing in the original space. We explored two types of latent spaces: one obtained from a dataset of randomized test scenarios and another from a dataset of optimized tests selected based on a known heuristic. The first latent space allows for the discovery of up to 71\% more failures compared to the original space, while also eliminating the need for custom search operators. The optimized latent space further improves the search process, leading to the discovery of 3.5 to 4 times more diverse failures compared to the original space.

These findings indicate that optimization in a latent space is an effective technique for improving search-based testing, particularly when a simple heuristic for identifying challenging test scenarios is available. The one-time computational overhead of RILaST can be reduced by leveraging parallelization techniques. As part of the future work, we plan to extend our approach to direct representation learning to further facilitate the search-based test generation process.

\begin{acks}
This research was supported by the Natural Sciences and Engineering Research Council of Canada (NSERC) [Grant No: RGPIN-2019-06956] and the Fonds de Recherche du Québec – Nature et technologies (FRQNT) [Grant No: 355318]. We also extend our gratitude to BeamNG.tech for providing an academic license for their simulator.
\end{acks}

\bibliographystyle{ACM-Reference-Format}
\bibliography{acmart}


\begin{thebibliography}{76}


\ifx \showCODEN    \undefined \def \showCODEN     #1{\unskip}     \fi
\ifx \showDOI      \undefined \def \showDOI       #1{#1}\fi
\ifx \showISBNx    \undefined \def \showISBNx     #1{\unskip}     \fi
\ifx \showISBNxiii \undefined \def \showISBNxiii  #1{\unskip}     \fi
\ifx \showISSN     \undefined \def \showISSN      #1{\unskip}     \fi
\ifx \showLCCN     \undefined \def \showLCCN      #1{\unskip}     \fi
\ifx \shownote     \undefined \def \shownote      #1{#1}          \fi
\ifx \showarticletitle \undefined \def \showarticletitle #1{#1}   \fi
\ifx \showURL      \undefined \def \showURL       {\relax}        \fi
\providecommand\bibfield[2]{#2}
\providecommand\bibinfo[2]{#2}
\providecommand\natexlab[1]{#1}
\providecommand\showeprint[2][]{arXiv:#2}

\bibitem[Adler and Lunz(2018)]%
        {adler2018banach}
\bibfield{author}{\bibinfo{person}{Jonas Adler} {and} \bibinfo{person}{Sebastian Lunz}.} \bibinfo{year}{2018}\natexlab{}.
\newblock \showarticletitle{Banach wasserstein gan}.
\newblock \bibinfo{journal}{\emph{Advances in neural information processing systems}}  \bibinfo{volume}{31} (\bibinfo{year}{2018}).
\newblock


\bibitem[Antoniol et~al\mbox{.}(2005)]%
        {antoniol2005search}
\bibfield{author}{\bibinfo{person}{Giuliano Antoniol}, \bibinfo{person}{Massimiliano Di~Penta}, {and} \bibinfo{person}{Mark Harman}.} \bibinfo{year}{2005}\natexlab{}.
\newblock \showarticletitle{Search-based techniques applied to optimization of project planning for a massive maintenance project}. In \bibinfo{booktitle}{\emph{21st IEEE International Conference on Software Maintenance (ICSM'05)}}. IEEE, \bibinfo{pages}{240--249}.
\newblock


\bibitem[Arcaini and Cetinkaya(2023)]%
        {arcaini2023crag}
\bibfield{author}{\bibinfo{person}{Paolo Arcaini} {and} \bibinfo{person}{Ahmet Cetinkaya}.} \bibinfo{year}{2023}\natexlab{}.
\newblock \showarticletitle{Crag at the sbft 2023 tool competition-cyber-physical systems track}. In \bibinfo{booktitle}{\emph{2023 IEEE/ACM International Workshop on Search-Based and Fuzz Testing (SBFT)}}. IEEE, \bibinfo{pages}{41--42}.
\newblock


\bibitem[Arcaini and Cetinkaya(2024)]%
        {arcaini2024crag}
\bibfield{author}{\bibinfo{person}{Paolo Arcaini} {and} \bibinfo{person}{Ahmet Cetinkaya}.} \bibinfo{year}{2024}\natexlab{}.
\newblock \showarticletitle{CRAG--a combinatorial testing-based generator of road geometries for ADS testing}.
\newblock \bibinfo{journal}{\emph{Science of Computer Programming}}  \bibinfo{volume}{238} (\bibinfo{year}{2024}), \bibinfo{pages}{103171}.
\newblock


\bibitem[Arcuri and Briand(2014)]%
        {arcuri2014hitchhiker}
\bibfield{author}{\bibinfo{person}{Andrea Arcuri} {and} \bibinfo{person}{Lionel Briand}.} \bibinfo{year}{2014}\natexlab{}.
\newblock \showarticletitle{A hitchhiker's guide to statistical tests for assessing randomized algorithms in software engineering}.
\newblock \bibinfo{journal}{\emph{Software Testing, Verification and Reliability}} \bibinfo{volume}{24}, \bibinfo{number}{3} (\bibinfo{year}{2014}), \bibinfo{pages}{219--250}.
\newblock


\bibitem[Arjovsky et~al\mbox{.}(2017)]%
        {arjovsky2017wasserstein}
\bibfield{author}{\bibinfo{person}{Martin Arjovsky}, \bibinfo{person}{Soumith Chintala}, {and} \bibinfo{person}{L{\'e}on Bottou}.} \bibinfo{year}{2017}\natexlab{}.
\newblock \showarticletitle{Wasserstein generative adversarial networks}. In \bibinfo{booktitle}{\emph{International conference on machine learning}}. PMLR, \bibinfo{pages}{214--223}.
\newblock


\bibitem[Babikian et~al\mbox{.}(2023)]%
        {babikian2023concretization}
\bibfield{author}{\bibinfo{person}{Aren~A Babikian}, \bibinfo{person}{Oszk{\'a}r Semer{\'a}th}, {and} \bibinfo{person}{D{\'a}niel Varr{\'o}}.} \bibinfo{year}{2023}\natexlab{}.
\newblock \showarticletitle{Concretization of Abstract Traffic Scene Specifications Using Metaheuristic Search}.
\newblock \bibinfo{journal}{\emph{IEEE Transactions on Software Engineering}} (\bibinfo{year}{2023}).
\newblock


\bibitem[B{\"a}ck et~al\mbox{.}(1997)]%
        {back1997handbook}
\bibfield{author}{\bibinfo{person}{Thomas B{\"a}ck}, \bibinfo{person}{David~B Fogel}, {and} \bibinfo{person}{Zbigniew Michalewicz}.} \bibinfo{year}{1997}\natexlab{}.
\newblock \showarticletitle{Handbook of evolutionary computation}.
\newblock \bibinfo{journal}{\emph{Release}} \bibinfo{volume}{97}, \bibinfo{number}{1} (\bibinfo{year}{1997}), \bibinfo{pages}{B1}.
\newblock


\bibitem[Bentley et~al\mbox{.}(2023)]%
        {bentley2023using}
\bibfield{author}{\bibinfo{person}{Peter Bentley}, \bibinfo{person}{Soo~Ling Lim}, \bibinfo{person}{Paolo Arcaini}, {and} \bibinfo{person}{Fuyuki Ishikawa}.} \bibinfo{year}{2023}\natexlab{}.
\newblock \showarticletitle{Using a variational autoencoder to learn valid search spaces of safely monitored autonomous robots for last-mile delivery}. In \bibinfo{booktitle}{\emph{Proceedings of the Genetic and Evolutionary Computation Conference}}. \bibinfo{pages}{1303--1311}.
\newblock


\bibitem[Bentley et~al\mbox{.}(2022a)]%
        {bentley2022coil}
\bibfield{author}{\bibinfo{person}{Peter~J Bentley}, \bibinfo{person}{Soo~Ling Lim}, \bibinfo{person}{Adam Gaier}, {and} \bibinfo{person}{Linh Tran}.} \bibinfo{year}{2022}\natexlab{a}.
\newblock \showarticletitle{COIL: Constrained optimization in learned latent space: learning representations for valid solutions}. In \bibinfo{booktitle}{\emph{Proceedings of the Genetic and Evolutionary Computation Conference Companion}}. \bibinfo{pages}{1870--1877}.
\newblock


\bibitem[Bentley et~al\mbox{.}(2022b)]%
        {bentley2022evolving}
\bibfield{author}{\bibinfo{person}{Peter~J Bentley}, \bibinfo{person}{Soo~Ling Lim}, \bibinfo{person}{Adam Gaier}, {and} \bibinfo{person}{Linh Tran}.} \bibinfo{year}{2022}\natexlab{b}.
\newblock \showarticletitle{Evolving through the looking glass: Learning improved search spaces with variational autoencoders}. In \bibinfo{booktitle}{\emph{International Conference on Parallel Problem Solving from Nature}}. Springer, \bibinfo{pages}{371--384}.
\newblock


\bibitem[Biagiola and Klikovits(2024)]%
        {10.1145/3643659.3643932}
\bibfield{author}{\bibinfo{person}{Matteo Biagiola} {and} \bibinfo{person}{Stefan Klikovits}.} \bibinfo{year}{2024}\natexlab{}.
\newblock \showarticletitle{SBFT Tool Competition 2024 - Cyber-Physical Systems Track}. In \bibinfo{booktitle}{\emph{Proceedings of the 17th ACM/IEEE International Workshop on Search-Based and Fuzz Testing}} (Lisbon, Portugal) \emph{(\bibinfo{series}{SBFT '24})}. \bibinfo{publisher}{Association for Computing Machinery}, \bibinfo{address}{New York, NY, USA}, \bibinfo{pages}{33–36}.
\newblock
\showISBNx{9798400705625}
\urldef\tempurl%
\url{https://doi.org/10.1145/3643659.3643932}
\showDOI{\tempurl}


\bibitem[Biagiola et~al\mbox{.}(2023)]%
        {biagiola2023sbft}
\bibfield{author}{\bibinfo{person}{Matteo Biagiola}, \bibinfo{person}{Stefan Klikovits}, \bibinfo{person}{Jarkko Peltom{\"a}ki}, {and} \bibinfo{person}{Vincenzo Riccio}.} \bibinfo{year}{2023}\natexlab{}.
\newblock \showarticletitle{Sbft tool competition 2023-cyber-physical systems track}. In \bibinfo{booktitle}{\emph{2023 IEEE/ACM International Workshop on Search-Based and Fuzz Testing (SBFT)}}. IEEE, \bibinfo{pages}{45--48}.
\newblock


\bibitem[Birchler et~al\mbox{.}(2022)]%
        {birchler2022cost}
\bibfield{author}{\bibinfo{person}{Christian Birchler}, \bibinfo{person}{Nicolas Ganz}, \bibinfo{person}{Sajad Khatiri}, \bibinfo{person}{Alessio Gambi}, {and} \bibinfo{person}{Sebastiano Panichella}.} \bibinfo{year}{2022}\natexlab{}.
\newblock \showarticletitle{Cost-effective simulation-based test selection in self-driving cars software with SDC-Scissor}. In \bibinfo{booktitle}{\emph{2022 IEEE International Conference on Software Analysis, Evolution and Reengineering (SANER)}}. IEEE, \bibinfo{pages}{164--168}.
\newblock


\bibitem[{Blank} and {Deb}(2020)]%
        {pymoo}
\bibfield{author}{\bibinfo{person}{J. {Blank}} {and} \bibinfo{person}{K. {Deb}}.} \bibinfo{year}{2020}\natexlab{}.
\newblock \showarticletitle{pymoo: Multi-Objective Optimization in Python}.
\newblock \bibinfo{journal}{\emph{IEEE Access}}  \bibinfo{volume}{8} (\bibinfo{year}{2020}), \bibinfo{pages}{89497--89509}.
\newblock


\bibitem[Bontrager et~al\mbox{.}(2018)]%
        {bontrager2018deep}
\bibfield{author}{\bibinfo{person}{Philip Bontrager}, \bibinfo{person}{Wending Lin}, \bibinfo{person}{Julian Togelius}, {and} \bibinfo{person}{Sebastian Risi}.} \bibinfo{year}{2018}\natexlab{}.
\newblock \showarticletitle{Deep interactive evolution}. In \bibinfo{booktitle}{\emph{Computational Intelligence in Music, Sound, Art and Design: 7th International Conference, EvoMUSART 2018, Parma, Italy, April 4-6, 2018, Proceedings}}. Springer, \bibinfo{pages}{267--282}.
\newblock


\bibitem[Campion et~al\mbox{.}(1996)]%
        {campion1996structural}
\bibfield{author}{\bibinfo{person}{Guy Campion}, \bibinfo{person}{Georges Bastin}, {and} \bibinfo{person}{Brigitte Dandrea-Novel}.} \bibinfo{year}{1996}\natexlab{}.
\newblock \showarticletitle{Structural properties and classification of kinematic and dynamic models of wheeled mobile robots}.
\newblock \bibinfo{journal}{\emph{IEEE transactions on robotics and automation}} \bibinfo{volume}{12}, \bibinfo{number}{1} (\bibinfo{year}{1996}), \bibinfo{pages}{47--62}.
\newblock


\bibitem[Castellano et~al\mbox{.}(2021)]%
        {castellano2021analysis}
\bibfield{author}{\bibinfo{person}{Ezequiel Castellano}, \bibinfo{person}{Ahmet Cetinkaya}, {and} \bibinfo{person}{Paolo Arcaini}.} \bibinfo{year}{2021}\natexlab{}.
\newblock \showarticletitle{Analysis of road representations in search-based testing of autonomous driving systems}. In \bibinfo{booktitle}{\emph{2021 IEEE 21st International Conference on Software Quality, Reliability and Security (QRS)}}. IEEE, \bibinfo{pages}{167--178}.
\newblock


\bibitem[Deb and Beyer(2001)]%
        {deb2001self}
\bibfield{author}{\bibinfo{person}{Kalyanmoy Deb} {and} \bibinfo{person}{Hans-Georg Beyer}.} \bibinfo{year}{2001}\natexlab{}.
\newblock \showarticletitle{Self-adaptive genetic algorithms with simulated binary crossover}.
\newblock \bibinfo{journal}{\emph{Evolutionary computation}} \bibinfo{volume}{9}, \bibinfo{number}{2} (\bibinfo{year}{2001}), \bibinfo{pages}{197--221}.
\newblock


\bibitem[Deb et~al\mbox{.}(2002)]%
        {deb2002fast}
\bibfield{author}{\bibinfo{person}{Kalyanmoy Deb}, \bibinfo{person}{Amrit Pratap}, \bibinfo{person}{Sameer Agarwal}, {and} \bibinfo{person}{TAMT Meyarivan}.} \bibinfo{year}{2002}\natexlab{}.
\newblock \showarticletitle{A fast and elitist multiobjective genetic algorithm: NSGA-II}.
\newblock \bibinfo{journal}{\emph{IEEE transactions on evolutionary computation}} \bibinfo{volume}{6}, \bibinfo{number}{2} (\bibinfo{year}{2002}), \bibinfo{pages}{182--197}.
\newblock


\bibitem[Deb et~al\mbox{.}(2007)]%
        {deb2007self}
\bibfield{author}{\bibinfo{person}{Kalyanmoy Deb}, \bibinfo{person}{Karthik Sindhya}, {and} \bibinfo{person}{Tatsuya Okabe}.} \bibinfo{year}{2007}\natexlab{}.
\newblock \showarticletitle{Self-adaptive simulated binary crossover for real-parameter optimization}. In \bibinfo{booktitle}{\emph{Proceedings of the 9th annual conference on genetic and evolutionary computation}}. \bibinfo{pages}{1187--1194}.
\newblock


\bibitem[Donz{\'e} and Maler(2010)]%
        {donze2010robust}
\bibfield{author}{\bibinfo{person}{Alexandre Donz{\'e}} {and} \bibinfo{person}{Oded Maler}.} \bibinfo{year}{2010}\natexlab{}.
\newblock \showarticletitle{Robust satisfaction of temporal logic over real-valued signals}. In \bibinfo{booktitle}{\emph{International Conference on Formal Modeling and Analysis of Timed Systems}}. Springer, \bibinfo{pages}{92--106}.
\newblock


\bibitem[Eiben and Smith(2015)]%
        {eiben2015introduction}
\bibfield{author}{\bibinfo{person}{Agoston~E Eiben} {and} \bibinfo{person}{James~E Smith}.} \bibinfo{year}{2015}\natexlab{}.
\newblock \bibinfo{booktitle}{\emph{Introduction to evolutionary computing}}.
\newblock \bibinfo{publisher}{Springer}.
\newblock


\bibitem[Fainekos and Pappas(2009)]%
        {fainekos2009robustness}
\bibfield{author}{\bibinfo{person}{Georgios~E Fainekos} {and} \bibinfo{person}{George~J Pappas}.} \bibinfo{year}{2009}\natexlab{}.
\newblock \showarticletitle{Robustness of temporal logic specifications for continuous-time signals}.
\newblock \bibinfo{journal}{\emph{Theoretical Computer Science}} \bibinfo{volume}{410}, \bibinfo{number}{42} (\bibinfo{year}{2009}), \bibinfo{pages}{4262--4291}.
\newblock


\bibitem[Foreman(2014)]%
        {foreman2014cosine}
\bibfield{author}{\bibinfo{person}{J Foreman}.} \bibinfo{year}{2014}\natexlab{}.
\newblock \bibinfo{title}{Cosine Distance, Cosine Similarity, Angular Cosine Distance, Angular Cosine Similarity}.
\newblock
\newblock


\bibitem[Gaier et~al\mbox{.}(2020)]%
        {gaier2020discovering}
\bibfield{author}{\bibinfo{person}{Adam Gaier}, \bibinfo{person}{Alexander Asteroth}, {and} \bibinfo{person}{Jean-Baptiste Mouret}.} \bibinfo{year}{2020}\natexlab{}.
\newblock \showarticletitle{Discovering representations for black-box optimization}. In \bibinfo{booktitle}{\emph{Proceedings of the 2020 Genetic and Evolutionary Computation Conference}}. \bibinfo{pages}{103--111}.
\newblock


\bibitem[Gambi et~al\mbox{.}(2022)]%
        {gambi2022sbst}
\bibfield{author}{\bibinfo{person}{Alessio Gambi}, \bibinfo{person}{Gunel Jahangirova}, \bibinfo{person}{Vincenzo Riccio}, {and} \bibinfo{person}{Fiorella Zampetti}.} \bibinfo{year}{2022}\natexlab{}.
\newblock \showarticletitle{SBST tool competition 2022}. In \bibinfo{booktitle}{\emph{2022 IEEE/ACM 15th International Workshop on Search-Based Software Testing (SBST)}}. IEEE, \bibinfo{pages}{25--32}.
\newblock


\bibitem[Gambi et~al\mbox{.}(2019)]%
        {gambi2019automatically}
\bibfield{author}{\bibinfo{person}{Alessio Gambi}, \bibinfo{person}{Marc Mueller}, {and} \bibinfo{person}{Gordon Fraser}.} \bibinfo{year}{2019}\natexlab{}.
\newblock \showarticletitle{Automatically testing self-driving cars with search-based procedural content generation}. In \bibinfo{booktitle}{\emph{Proceedings of the 28th ACM SIGSOFT International Symposium on Software Testing and Analysis}}. \bibinfo{pages}{318--328}.
\newblock


\bibitem[GmbH(2021)]%
        {beamng}
\bibfield{author}{\bibinfo{person}{BeamNG GmbH}.} \bibinfo{year}{2021}\natexlab{}.
\newblock \bibinfo{title}{{BeamNGpy}}.
\newblock \bibinfo{howpublished}{\url{https://github.com/BeamNG/BeamNGpy}}.
\newblock


\bibitem[Gog et~al\mbox{.}(2021)]%
        {gog2021pylot}
\bibfield{author}{\bibinfo{person}{Ionel Gog}, \bibinfo{person}{Sukrit Kalra}, \bibinfo{person}{Peter Schafhalter}, \bibinfo{person}{Matthew~A Wright}, \bibinfo{person}{Joseph~E Gonzalez}, {and} \bibinfo{person}{Ion Stoica}.} \bibinfo{year}{2021}\natexlab{}.
\newblock \showarticletitle{Pylot: A modular platform for exploring latency-accuracy tradeoffs in autonomous vehicles}. In \bibinfo{booktitle}{\emph{2021 IEEE International Conference on Robotics and Automation (ICRA)}}. IEEE, \bibinfo{pages}{8806--8813}.
\newblock


\bibitem[Goldberg and Deb(1991)]%
        {goldberg1991comparative}
\bibfield{author}{\bibinfo{person}{David~E Goldberg} {and} \bibinfo{person}{Kalyanmoy Deb}.} \bibinfo{year}{1991}\natexlab{}.
\newblock \showarticletitle{A comparative analysis of selection schemes used in genetic algorithms}.
\newblock In \bibinfo{booktitle}{\emph{Foundations of genetic algorithms}}. Vol.~\bibinfo{volume}{1}. \bibinfo{publisher}{Elsevier}, \bibinfo{pages}{69--93}.
\newblock


\bibitem[Goodfellow et~al\mbox{.}(2016)]%
        {Goodfellow-et-al-2016}
\bibfield{author}{\bibinfo{person}{Ian Goodfellow}, \bibinfo{person}{Yoshua Bengio}, {and} \bibinfo{person}{Aaron Courville}.} \bibinfo{year}{2016}\natexlab{}.
\newblock \bibinfo{booktitle}{\emph{Deep Learning}}.
\newblock \bibinfo{publisher}{MIT Press}.
\newblock
\newblock
\shownote{\url{http://www.deeplearningbook.org}}.


\bibitem[Goodfellow et~al\mbox{.}(2014)]%
        {goodfellow2014generative}
\bibfield{author}{\bibinfo{person}{Ian Goodfellow}, \bibinfo{person}{Jean Pouget-Abadie}, \bibinfo{person}{Mehdi Mirza}, \bibinfo{person}{Bing Xu}, \bibinfo{person}{David Warde-Farley}, \bibinfo{person}{Sherjil Ozair}, \bibinfo{person}{Aaron Courville}, {and} \bibinfo{person}{Yoshua Bengio}.} \bibinfo{year}{2014}\natexlab{}.
\newblock \showarticletitle{Generative adversarial nets}.
\newblock \bibinfo{journal}{\emph{Advances in neural information processing systems}}  \bibinfo{volume}{27} (\bibinfo{year}{2014}).
\newblock


\bibitem[Han et~al\mbox{.}(2022)]%
        {han2022data}
\bibfield{author}{\bibinfo{person}{Jiawei Han}, \bibinfo{person}{Jian Pei}, {and} \bibinfo{person}{Hanghang Tong}.} \bibinfo{year}{2022}\natexlab{}.
\newblock \bibinfo{booktitle}{\emph{Data mining: concepts and techniques}}.
\newblock \bibinfo{publisher}{Morgan kaufmann}.
\newblock


\bibitem[Haq et~al\mbox{.}(2022)]%
        {haq2022efficient}
\bibfield{author}{\bibinfo{person}{Fitash~Ul Haq}, \bibinfo{person}{Donghwan Shin}, {and} \bibinfo{person}{Lionel Briand}.} \bibinfo{year}{2022}\natexlab{}.
\newblock \showarticletitle{Efficient online testing for DNN-enabled systems using surrogate-assisted and many-objective optimization}. In \bibinfo{booktitle}{\emph{Proceedings of the 44th international conference on software engineering}}. \bibinfo{pages}{811--822}.
\newblock


\bibitem[Harman et~al\mbox{.}(2008)]%
        {harman2008search}
\bibfield{author}{\bibinfo{person}{Mark Harman}, \bibinfo{person}{Phil McMinn}, \bibinfo{person}{Jerffeson~Teixeira De~Souza}, {and} \bibinfo{person}{Shin Yoo}.} \bibinfo{year}{2008}\natexlab{}.
\newblock \showarticletitle{Search based software engineering: Techniques, taxonomy, tutorial}.
\newblock In \bibinfo{booktitle}{\emph{LASER Summer School on Software Engineering}}. \bibinfo{publisher}{Springer}, \bibinfo{pages}{1--59}.
\newblock


\bibitem[Huai et~al\mbox{.}(2023a)]%
        {huai2023sceno}
\bibfield{author}{\bibinfo{person}{Yuqi Huai}, \bibinfo{person}{Sumaya Almanee}, \bibinfo{person}{Yuntianyi Chen}, \bibinfo{person}{Xiafa Wu}, \bibinfo{person}{Qi~Alfred Chen}, {and} \bibinfo{person}{Joshua Garcia}.} \bibinfo{year}{2023}\natexlab{a}.
\newblock \showarticletitle{sceno RITA: Generating Diverse, Fully-Mutable, Test Scenarios for Autonomous Vehicle Planning}.
\newblock \bibinfo{journal}{\emph{IEEE Transactions on Software Engineering}} (\bibinfo{year}{2023}).
\newblock


\bibitem[Huai et~al\mbox{.}(2023b)]%
        {huai2023doppelganger}
\bibfield{author}{\bibinfo{person}{Yuqi Huai}, \bibinfo{person}{Yuntianyi Chen}, \bibinfo{person}{Sumaya Almanee}, \bibinfo{person}{Tuan Ngo}, \bibinfo{person}{Xiang Liao}, \bibinfo{person}{Ziwen Wan}, \bibinfo{person}{Qi~Alfred Chen}, {and} \bibinfo{person}{Joshua Garcia}.} \bibinfo{year}{2023}\natexlab{b}.
\newblock \showarticletitle{Doppelg{\"a}nger Test Generation for Revealing Bugs in Autonomous Driving Software}. In \bibinfo{booktitle}{\emph{2023 IEEE/ACM 45th International Conference on Software Engineering (ICSE)}}. IEEE, \bibinfo{pages}{2591--2603}.
\newblock


\bibitem[Humeniuk and Khomh(2024)]%
        {humeniuk2024ambiegen}
\bibfield{author}{\bibinfo{person}{Dmytro Humeniuk} {and} \bibinfo{person}{Foutse Khomh}.} \bibinfo{year}{2024}\natexlab{}.
\newblock \showarticletitle{AmbieGen at the SBFT 2024 Tool Competition-CPS-UAV Track}. In \bibinfo{booktitle}{\emph{Proceedings of the 17th ACM/IEEE International Workshop on Search-Based and Fuzz Testing}}. \bibinfo{pages}{69--70}.
\newblock


\bibitem[Humeniuk and Khomh(2025)]%
        {humeniuk_2025_15087028}
\bibfield{author}{\bibinfo{person}{Dmytro Humeniuk} {and} \bibinfo{person}{Foutse Khomh}.} \bibinfo{year}{2025}\natexlab{}.
\newblock \bibinfo{booktitle}{\emph{Representation Improvement in Latent Space for Search Based Testing of Autonomous Robotic Systems - replication package}}.
\newblock
\urldef\tempurl%
\url{https://doi.org/10.5281/zenodo.15087028}
\showDOI{\tempurl}


\bibitem[Humeniuk et~al\mbox{.}(2022)]%
        {humeniuk2022search}
\bibfield{author}{\bibinfo{person}{Dmytro Humeniuk}, \bibinfo{person}{Foutse Khomh}, {and} \bibinfo{person}{Giuliano Antoniol}.} \bibinfo{year}{2022}\natexlab{}.
\newblock \showarticletitle{A search-based framework for automatic generation of testing environments for cyber--physical systems}.
\newblock \bibinfo{journal}{\emph{Information and Software Technology}}  \bibinfo{volume}{149} (\bibinfo{year}{2022}), \bibinfo{pages}{106936}.
\newblock


\bibitem[Humeniuk et~al\mbox{.}(2023)]%
        {humeniuk2023ambiegen}
\bibfield{author}{\bibinfo{person}{Dmytro Humeniuk}, \bibinfo{person}{Foutse Khomh}, {and} \bibinfo{person}{Giuliano Antoniol}.} \bibinfo{year}{2023}\natexlab{}.
\newblock \showarticletitle{AmbieGen: A Search-based Framework for Autonomous Systems Testing}.
\newblock \bibinfo{journal}{\emph{arXiv preprint arXiv:2301.01234}} (\bibinfo{year}{2023}).
\newblock


\bibitem[Humeniuk et~al\mbox{.}(2024)]%
        {rigaa}
\bibfield{author}{\bibinfo{person}{Dmytro Humeniuk}, \bibinfo{person}{Foutse Khomh}, {and} \bibinfo{person}{Giuliano Antoniol}.} \bibinfo{year}{2024}\natexlab{}.
\newblock \showarticletitle{Reinforcement Learning Informed Evolutionary Search for Autonomous Systems Testing}.
\newblock \bibinfo{journal}{\emph{ACM Trans. Softw. Eng. Methodol.}} (\bibinfo{date}{jul} \bibinfo{year}{2024}).
\newblock
\showISSN{1049-331X}
\urldef\tempurl%
\url{https://doi.org/10.1145/3680468}
\showDOI{\tempurl}
\newblock
\shownote{Just Accepted}.


\bibitem[Khatiri et~al\mbox{.}(2023)]%
        {khatiri2023simulation}
\bibfield{author}{\bibinfo{person}{Sajad Khatiri}, \bibinfo{person}{Sebastiano Panichella}, {and} \bibinfo{person}{Paolo Tonella}.} \bibinfo{year}{2023}\natexlab{}.
\newblock \showarticletitle{Simulation-based test case generation for unmanned aerial vehicles in the neighborhood of real flights}. In \bibinfo{booktitle}{\emph{2023 IEEE Conference on Software Testing, Verification and Validation (ICST)}}. IEEE, \bibinfo{pages}{281--292}.
\newblock


\bibitem[Khatiri et~al\mbox{.}(2024a)]%
        {khatiri2024simulation}
\bibfield{author}{\bibinfo{person}{Sajad Khatiri}, \bibinfo{person}{Sebastiano Panichella}, {and} \bibinfo{person}{Paolo Tonella}.} \bibinfo{year}{2024}\natexlab{a}.
\newblock \showarticletitle{Simulation-based testing of unmanned aerial vehicles with aerialist}. In \bibinfo{booktitle}{\emph{Proceedings of the 2024 IEEE/ACM 46th International Conference on Software Engineering: Companion Proceedings}}. \bibinfo{pages}{134--138}.
\newblock


\bibitem[Khatiri et~al\mbox{.}(2024b)]%
        {icse2024Aerialist}
\bibfield{author}{\bibinfo{person}{Sajad Khatiri}, \bibinfo{person}{Sebastiano Panichella}, {and} \bibinfo{person}{Paolo Tonella}.} \bibinfo{year}{2024}\natexlab{b}.
\newblock \showarticletitle{Simulation-based Testing of Unmanned Aerial Vehicles with Aerialist}. In \bibinfo{booktitle}{\emph{International Conference on Software Engineering (ICSE)}}.
\newblock


\bibitem[Khatiri et~al\mbox{.}(2024c)]%
        {uav-competition}
\bibfield{author}{\bibinfo{person}{Sajad Khatiri}, \bibinfo{person}{Prasun Saurabh}, \bibinfo{person}{Timothy Zimmermann}, \bibinfo{person}{Charith Munasinghe}, \bibinfo{person}{Christian Birchler}, {and} \bibinfo{person}{Sebastiano Panichella}.} \bibinfo{year}{2024}\natexlab{c}.
\newblock \showarticletitle{SBFT Tool Competition 2024 - CPS-UAV Test Case Generation Track}. In \bibinfo{booktitle}{\emph{Proceedings of the 17th ACM/IEEE International Workshop on Search-Based and Fuzz Testing}} (Lisbon, Portugal) \emph{(\bibinfo{series}{SBFT '24})}. \bibinfo{publisher}{Association for Computing Machinery}, \bibinfo{address}{New York, NY, USA}, \bibinfo{pages}{29–32}.
\newblock
\showISBNx{9798400705625}
\urldef\tempurl%
\url{https://doi.org/10.1145/3643659.3643931}
\showDOI{\tempurl}


\bibitem[Khatiri et~al\mbox{.}(2024d)]%
        {khatiri2024sbft}
\bibfield{author}{\bibinfo{person}{Sajad Khatiri}, \bibinfo{person}{Prasun Saurabh}, \bibinfo{person}{Timothy Zimmermann}, \bibinfo{person}{Charith Munasinghe}, \bibinfo{person}{Christian Birchler}, {and} \bibinfo{person}{Sebastiano Panichella}.} \bibinfo{year}{2024}\natexlab{d}.
\newblock \showarticletitle{SBFT tool competition 2024: CPS-UAV test case generation track}. In \bibinfo{booktitle}{\emph{17th International Workshop on Search-Based and Fuzz Testing (SBFT), Lisbon, Portugal, 14-20 April 2024}}. ZHAW Z{\"u}rcher Hochschule f{\"u}r Angewandte Wissenschaften.
\newblock


\bibitem[Kingma(2013)]%
        {kingma2013auto}
\bibfield{author}{\bibinfo{person}{Diederik~P Kingma}.} \bibinfo{year}{2013}\natexlab{}.
\newblock \showarticletitle{Auto-encoding variational bayes}.
\newblock \bibinfo{journal}{\emph{arXiv preprint arXiv:1312.6114}} (\bibinfo{year}{2013}).
\newblock


\bibitem[Kingma(2014)]%
        {kingma2014adam}
\bibfield{author}{\bibinfo{person}{Diederik~P Kingma}.} \bibinfo{year}{2014}\natexlab{}.
\newblock \showarticletitle{Adam: A method for stochastic optimization}.
\newblock \bibinfo{journal}{\emph{arXiv preprint arXiv:1412.6980}} (\bibinfo{year}{2014}).
\newblock


\bibitem[Kingma et~al\mbox{.}(2019)]%
        {kingma2019introduction}
\bibfield{author}{\bibinfo{person}{Diederik~P Kingma}, \bibinfo{person}{Max Welling}, {et~al\mbox{.}}} \bibinfo{year}{2019}\natexlab{}.
\newblock \showarticletitle{An introduction to variational autoencoders}.
\newblock \bibinfo{journal}{\emph{Foundations and Trends{\textregistered} in Machine Learning}} \bibinfo{volume}{12}, \bibinfo{number}{4} (\bibinfo{year}{2019}), \bibinfo{pages}{307--392}.
\newblock


\bibitem[Klikovits et~al\mbox{.}(2023)]%
        {klikovits2023frenetic}
\bibfield{author}{\bibinfo{person}{Stefan Klikovits}, \bibinfo{person}{Ezequiel Castellano}, \bibinfo{person}{Ahmet Cetinkaya}, {and} \bibinfo{person}{Paolo Arcaini}.} \bibinfo{year}{2023}\natexlab{}.
\newblock \showarticletitle{Frenetic-lib: An extensible framework for search-based generation of road structures for ADS testing}.
\newblock \bibinfo{journal}{\emph{Science of Computer Programming}}  \bibinfo{volume}{230} (\bibinfo{year}{2023}), \bibinfo{pages}{102996}.
\newblock


\bibitem[Koenig and Howard(2004)]%
        {1389727}
\bibfield{author}{\bibinfo{person}{N. Koenig} {and} \bibinfo{person}{A. Howard}.} \bibinfo{year}{2004}\natexlab{}.
\newblock \showarticletitle{Design and use paradigms for Gazebo, an open-source multi-robot simulator}. In \bibinfo{booktitle}{\emph{2004 IEEE/RSJ International Conference on Intelligent Robots and Systems (IROS) (IEEE Cat. No.04CH37566)}}, Vol.~\bibinfo{volume}{3}. \bibinfo{pages}{2149--2154 vol.3}.
\newblock
\urldef\tempurl%
\url{https://doi.org/10.1109/IROS.2004.1389727}
\showDOI{\tempurl}


\bibitem[LaValle(1998)]%
        {lavalle1998rapidly}
\bibfield{author}{\bibinfo{person}{Steven LaValle}.} \bibinfo{year}{1998}\natexlab{}.
\newblock \showarticletitle{Rapidly-exploring random trees: A new tool for path planning}.
\newblock \bibinfo{journal}{\emph{Research Report 9811}} (\bibinfo{year}{1998}).
\newblock


\bibitem[Lehman et~al\mbox{.}(2008)]%
        {lehman2008exploiting}
\bibfield{author}{\bibinfo{person}{Joel Lehman}, \bibinfo{person}{Kenneth~O Stanley}, {et~al\mbox{.}}} \bibinfo{year}{2008}\natexlab{}.
\newblock \showarticletitle{Exploiting open-endedness to solve problems through the search for novelty.}. In \bibinfo{booktitle}{\emph{ALIFE}}. \bibinfo{pages}{329--336}.
\newblock


\bibitem[Levenshtein et~al\mbox{.}(1966)]%
        {levenshtein1966binary}
\bibfield{author}{\bibinfo{person}{Vladimir~I Levenshtein} {et~al\mbox{.}}} \bibinfo{year}{1966}\natexlab{}.
\newblock \showarticletitle{Binary codes capable of correcting deletions, insertions, and reversals}. In \bibinfo{booktitle}{\emph{Soviet physics doklady}}, Vol.~\bibinfo{volume}{10}. Soviet Union, \bibinfo{pages}{707--710}.
\newblock


\bibitem[Lu et~al\mbox{.}(2024)]%
        {lu2024epitester}
\bibfield{author}{\bibinfo{person}{Chengjie Lu}, \bibinfo{person}{Shaukat Ali}, {and} \bibinfo{person}{Tao Yue}.} \bibinfo{year}{2024}\natexlab{}.
\newblock \showarticletitle{Epitester: Testing autonomous vehicles with epigenetic algorithm and attention mechanism}.
\newblock \bibinfo{journal}{\emph{IEEE Transactions on Software Engineering}} (\bibinfo{year}{2024}).
\newblock


\bibitem[Meier et~al\mbox{.}(2015)]%
        {meier2015px4}
\bibfield{author}{\bibinfo{person}{Lorenz Meier}, \bibinfo{person}{Dominik Honegger}, {and} \bibinfo{person}{Marc Pollefeys}.} \bibinfo{year}{2015}\natexlab{}.
\newblock \showarticletitle{PX4: A node-based multithreaded open source robotics framework for deeply embedded platforms}. In \bibinfo{booktitle}{\emph{2015 IEEE international conference on robotics and automation (ICRA)}}. IEEE, \bibinfo{pages}{6235--6240}.
\newblock


\bibitem[Menghi et~al\mbox{.}(2020)]%
        {menghi2020approximation}
\bibfield{author}{\bibinfo{person}{Claudio Menghi}, \bibinfo{person}{Shiva Nejati}, \bibinfo{person}{Lionel Briand}, {and} \bibinfo{person}{Yago~Isasi Parache}.} \bibinfo{year}{2020}\natexlab{}.
\newblock \showarticletitle{Approximation-refinement testing of compute-intensive cyber-physical models: An approach based on system identification}. In \bibinfo{booktitle}{\emph{Proceedings of the ACM/IEEE 42nd International Conference on Software Engineering}}. \bibinfo{pages}{372--384}.
\newblock


\bibitem[Mouret and Clune(2015)]%
        {mouret2015illuminating}
\bibfield{author}{\bibinfo{person}{Jean-Baptiste Mouret} {and} \bibinfo{person}{Jeff Clune}.} \bibinfo{year}{2015}\natexlab{}.
\newblock \showarticletitle{Illuminating search spaces by mapping elites}.
\newblock \bibinfo{journal}{\emph{arXiv preprint arXiv:1504.04909}} (\bibinfo{year}{2015}).
\newblock


\bibitem[Panichella et~al\mbox{.}(2021)]%
        {panichella2021sbst}
\bibfield{author}{\bibinfo{person}{Sebastiano Panichella}, \bibinfo{person}{Alessio Gambi}, \bibinfo{person}{Fiorella Zampetti}, {and} \bibinfo{person}{Vincenzo Riccio}.} \bibinfo{year}{2021}\natexlab{}.
\newblock \showarticletitle{Sbst tool competition 2021}. In \bibinfo{booktitle}{\emph{2021 IEEE/ACM 14th International Workshop on Search-Based Software Testing (SBST)}}. IEEE, \bibinfo{pages}{20--27}.
\newblock


\bibitem[Paszke et~al\mbox{.}(2019)]%
        {paszke2019pytorch}
\bibfield{author}{\bibinfo{person}{Adam Paszke}, \bibinfo{person}{Sam Gross}, \bibinfo{person}{Francisco Massa}, \bibinfo{person}{Adam Lerer}, \bibinfo{person}{James Bradbury}, \bibinfo{person}{Gregory Chanan}, \bibinfo{person}{Trevor Killeen}, \bibinfo{person}{Zeming Lin}, \bibinfo{person}{Natalia Gimelshein}, \bibinfo{person}{Luca Antiga}, {et~al\mbox{.}}} \bibinfo{year}{2019}\natexlab{}.
\newblock \showarticletitle{Pytorch: An imperative style, high-performance deep learning library}.
\newblock \bibinfo{journal}{\emph{Advances in neural information processing systems}}  \bibinfo{volume}{32} (\bibinfo{year}{2019}).
\newblock


\bibitem[Peltom{\"a}ki and Porres(2024)]%
        {peltomaki2024learning}
\bibfield{author}{\bibinfo{person}{Jarkko Peltom{\"a}ki} {and} \bibinfo{person}{Ivan Porres}.} \bibinfo{year}{2024}\natexlab{}.
\newblock \showarticletitle{Learning test generators for cyber-physical systems}.
\newblock \bibinfo{journal}{\emph{arXiv preprint arXiv:2410.03202}} (\bibinfo{year}{2024}).
\newblock


\bibitem[Riccio and Tonella(2020)]%
        {riccio2020model}
\bibfield{author}{\bibinfo{person}{Vincenzo Riccio} {and} \bibinfo{person}{Paolo Tonella}.} \bibinfo{year}{2020}\natexlab{}.
\newblock \showarticletitle{Model-based exploration of the frontier of behaviours for deep learning system testing}. In \bibinfo{booktitle}{\emph{Proceedings of the 28th ACM Joint Meeting on European Software Engineering Conference and Symposium on the Foundations of Software Engineering}}. \bibinfo{pages}{876--888}.
\newblock


\bibitem[Rothlauf and Rothlauf(2002)]%
        {rothlauf2002representations}
\bibfield{author}{\bibinfo{person}{Franz Rothlauf} {and} \bibinfo{person}{Franz Rothlauf}.} \bibinfo{year}{2002}\natexlab{}.
\newblock \bibinfo{booktitle}{\emph{Representations for genetic and evolutionary algorithms}}.
\newblock \bibinfo{publisher}{Springer}.
\newblock


\bibitem[Smith et~al\mbox{.}(2017)]%
        {smith2017don}
\bibfield{author}{\bibinfo{person}{Samuel~L Smith}, \bibinfo{person}{Pieter-Jan Kindermans}, \bibinfo{person}{Chris Ying}, {and} \bibinfo{person}{Quoc~V Le}.} \bibinfo{year}{2017}\natexlab{}.
\newblock \showarticletitle{Don't decay the learning rate, increase the batch size}.
\newblock \bibinfo{journal}{\emph{arXiv preprint arXiv:1711.00489}} (\bibinfo{year}{2017}).
\newblock


\bibitem[Sorokin and Kerscher(2024)]%
        {sorokin2024guiding}
\bibfield{author}{\bibinfo{person}{Lev Sorokin} {and} \bibinfo{person}{Niklas Kerscher}.} \bibinfo{year}{2024}\natexlab{}.
\newblock \showarticletitle{Guiding the search towards failure-inducing test inputs using support vector machines}. In \bibinfo{booktitle}{\emph{Proceedings of the 5th IEEE/ACM International Workshop on Deep Learning for Testing and Testing for Deep Learning}}. \bibinfo{pages}{9--12}.
\newblock


\bibitem[Stolfi and Alba(2018)]%
        {epiga}
\bibfield{author}{\bibinfo{person}{Daniel~H. Stolfi} {and} \bibinfo{person}{Enrique Alba}.} \bibinfo{year}{2018}\natexlab{}.
\newblock \showarticletitle{Epigenetic algorithms: A New way of building GAs based on epigenetics}.
\newblock \bibinfo{journal}{\emph{Inf. Sci.}} \bibinfo{volume}{424}, \bibinfo{number}{C} (\bibinfo{date}{Jan.} \bibinfo{year}{2018}), \bibinfo{pages}{250–272}.
\newblock
\showISSN{0020-0255}
\urldef\tempurl%
\url{https://doi.org/10.1016/j.ins.2017.10.005}
\showDOI{\tempurl}


\bibitem[Tang et~al\mbox{.}(2024)]%
        {tang2024tumb}
\bibfield{author}{\bibinfo{person}{Shuncheng Tang}, \bibinfo{person}{Zhenya Zhang}, \bibinfo{person}{Ahmet Cetinkaya}, {and} \bibinfo{person}{Paolo Arcaini}.} \bibinfo{year}{2024}\natexlab{}.
\newblock \showarticletitle{TUMB at the SBFT 2024 Tool Competition-CPS-UAV Test Case Generation Track}. In \bibinfo{booktitle}{\emph{Proceedings of the 17th ACM/IEEE International Workshop on Search-Based and Fuzz Testing}}. \bibinfo{pages}{53--54}.
\newblock


\bibitem[Tapp(2016)]%
        {tapp2016differential}
\bibfield{author}{\bibinfo{person}{Kristopher Tapp}.} \bibinfo{year}{2016}\natexlab{}.
\newblock \bibinfo{booktitle}{\emph{Differential geometry of curves and surfaces}}.
\newblock \bibinfo{publisher}{Springer}.
\newblock


\bibitem[Tuncali et~al\mbox{.}(2018)]%
        {tuncali2018simulation}
\bibfield{author}{\bibinfo{person}{Cumhur~Erkan Tuncali}, \bibinfo{person}{Georgios Fainekos}, \bibinfo{person}{Hisahiro Ito}, {and} \bibinfo{person}{James Kapinski}.} \bibinfo{year}{2018}\natexlab{}.
\newblock \showarticletitle{Simulation-based adversarial test generation for autonomous vehicles with machine learning components}. In \bibinfo{booktitle}{\emph{2018 IEEE Intelligent Vehicles Symposium (IV)}}. IEEE, \bibinfo{pages}{1555--1562}.
\newblock


\bibitem[Umbarkar and Sheth(2015)]%
        {umbarkar2015crossover}
\bibfield{author}{\bibinfo{person}{Anant~J Umbarkar} {and} \bibinfo{person}{Pranali~D Sheth}.} \bibinfo{year}{2015}\natexlab{}.
\newblock \showarticletitle{Crossover operators in genetic algorithms: a review.}
\newblock \bibinfo{journal}{\emph{ICTACT journal on soft computing}} \bibinfo{volume}{6}, \bibinfo{number}{1} (\bibinfo{year}{2015}).
\newblock


\bibitem[Volz et~al\mbox{.}(2018)]%
        {volz2018evolving}
\bibfield{author}{\bibinfo{person}{Vanessa Volz}, \bibinfo{person}{Jacob Schrum}, \bibinfo{person}{Jialin Liu}, \bibinfo{person}{Simon~M Lucas}, \bibinfo{person}{Adam Smith}, {and} \bibinfo{person}{Sebastian Risi}.} \bibinfo{year}{2018}\natexlab{}.
\newblock \showarticletitle{Evolving mario levels in the latent space of a deep convolutional generative adversarial network}. In \bibinfo{booktitle}{\emph{Proceedings of the genetic and evolutionary computation conference}}. \bibinfo{pages}{221--228}.
\newblock


\bibitem[Winsten et~al\mbox{.}(2024)]%
        {winsten2024adaptive}
\bibfield{author}{\bibinfo{person}{Jesper Winsten}, \bibinfo{person}{Valentin Soloviev}, \bibinfo{person}{Jarkko Peltom{\"a}ki}, {and} \bibinfo{person}{Ivan Porres}.} \bibinfo{year}{2024}\natexlab{}.
\newblock \showarticletitle{Adaptive test generation for unmanned aerial vehicles using WOGAN-UAV}. In \bibinfo{booktitle}{\emph{Proceedings of the 17th ACM/IEEE International Workshop on Search-Based and Fuzz Testing}}. \bibinfo{pages}{43--44}.
\newblock


\bibitem[Zhang and Li(2007)]%
        {zhang2007moea}
\bibfield{author}{\bibinfo{person}{Qingfu Zhang} {and} \bibinfo{person}{Hui Li}.} \bibinfo{year}{2007}\natexlab{}.
\newblock \showarticletitle{MOEA/D: A multiobjective evolutionary algorithm based on decomposition}.
\newblock \bibinfo{journal}{\emph{IEEE Transactions on evolutionary computation}} \bibinfo{volume}{11}, \bibinfo{number}{6} (\bibinfo{year}{2007}), \bibinfo{pages}{712--731}.
\newblock


\bibitem[Zohdinasab et~al\mbox{.}(2024)]%
        {zohdinasab2024focused}
\bibfield{author}{\bibinfo{person}{Tahereh Zohdinasab}, \bibinfo{person}{Vincenzo Riccio}, {and} \bibinfo{person}{Paolo Tonella}.} \bibinfo{year}{2024}\natexlab{}.
\newblock \showarticletitle{Focused Test Generation for Autonomous Driving Systems}.
\newblock \bibinfo{journal}{\emph{ACM Transactions on Software Engineering and Methodology}} (\bibinfo{year}{2024}).
\newblock


\end{thebibliography}

\appendix
\section{Results of statistical significance testing for RQ2} \label{sec:appendix}
\begin{table}[!ht]
    \centering
    \caption{Number of revealed failures and their sparseness in UC1}
    \scalebox{0.9}{
    \begin{tabular}{cccc|cccc}
    \hline
        Number of failures & ~ & ~ & ~ & Sparseness & ~ & ~ & ~ \\ \hline
        A & B & p-value & Effect size & A & B & p-value & Effect size \\ 
        ran & ga1 & 0.025 & -0.6, L & ran & ga1 & 0.97 & -0.02, N \\ 
        ran & ga2 & 0.001 & -0.89, L & ran & ga2 & 0.076 & 0.48, L \\ 
        ran & vr\_ran & 0.005 & 0.71, L & ran & vr\_ran & 0.427 & -0.22, S \\ 
        ran & vr\_ga1 & 0.444 & -0.21, S & ran & vr\_ga1 & 0.734 & 0.1, N \\ 
        ran & vr\_ga2 & 0 & -1.0, L & ran & vr\_ga2 & 0.93 & 0.029, N \\ 
        ran & vo\_ran & 0 & -1.0, L & ran & vo\_ran & 0.241 & 0.32, S \\ 
        ran & vo\_ga1 & 0 & -1.0, L & ran & vo\_ga1 & 0.006 & 0.74, L \\ 
        ran & vo\_ga2 & 0 & -1.0, L & ran & vo\_ga2 & 0.254 & 0.286, S \\ 
        ga1 & ga2 & 0.288 & -0.29, S & ga1 & ga2 & 0.031 & 0.58, L \\ 
        ga1 & vr\_ran & 0.001 & 0.87, L & ga1 & vr\_ran & 0.385 & -0.24, S \\ 
        ga1 & vr\_ga1 & 0.149 & 0.39, M & ga1 & vr\_ga1 & 0.385 & 0.24, S \\ 
        ga1 & vr\_ga2 & 0.168 & -0.343, M & ga1 & vr\_ga2 & 0.539 & 0.157, S \\ 
        ga1 & vo\_ran & 0.001 & -0.91, L & ga1 & vo\_ran & 0.021 & 0.62, L \\ 
        ga1 & vo\_ga1 & 0.001 & -0.91, L & ga1 & vo\_ga1 & 0.003 & 0.8, L \\ 
        ga1 & vo\_ga2 & 0 & -1.0, L & ga1 & vo\_ga2 & 0.05 & 0.486, L \\ 
        ga2 & vr\_ran & 0 & 1.0, L & ga2 & vr\_ran & 0.089 & -0.46, M \\ 
        ga2 & vr\_ga1 & 0.009 & 0.7, L & ga2 & vr\_ga1 & 0.089 & -0.46, M \\ 
        ga2 & vr\_ga2 & 0.953 & -0.021, N & ga2 & vr\_ga2 & 0.018 & -0.586, L \\ 
        ga2 & vo\_ran & 0.001 & -0.91, L & ga2 & vo\_ran & 0.186 & -0.36, M \\ 
        ga2 & vo\_ga1 & 0.001 & -0.86, L & ga2 & vo\_ga1 & 0.97 & -0.02, N \\ 
        ga2 & vo\_ga2 & 0 & -1.0, L & ga2 & vo\_ga2 & 0.135 & -0.371, M \\ 
        vr\_ran & vr\_ga1 & 0.002 & -0.81, L & vr\_ran & vr\_ga1 & 0.273 & 0.3, S \\ 
        vr\_ran & vr\_ga2 & 0 & -1.0, L & vr\_ran & vr\_ga2 & 0.364 & 0.229, S \\ 
        vr\_ran & vo\_ran & 0 & -1.0, L & vr\_ran & vo\_ran & 0.273 & 0.3, S \\ 
        vr\_ran & vo\_ga1 & 0 & -1.0, L & vr\_ran & vo\_ga1 & 0.089 & 0.46, M \\ 
        vr\_ran & vo\_ga2 & 0 & -1.0, L & vr\_ran & vo\_ga2 & 0.23 & 0.3, S \\ 
        vr\_ga1 & vr\_ga2 & 0.004 & -0.714, L & vr\_ga1 & vr\_ga2 & 0.93 & -0.029, N \\ 
        vr\_ga1 & vo\_ran & 0 & -1.0, L & vr\_ga1 & vo\_ran & 0.064 & 0.5, L \\ 
        vr\_ga1 & vo\_ga1 & 0 & -0.96, L & vr\_ga1 & vo\_ga1 & 0.011 & 0.68, L \\ 
        vr\_ga1 & vo\_ga2 & 0 & -1.0, L & vr\_ga1 & vo\_ga2 & 0.188 & 0.329, S \\ 
        vr\_ga2 & vo\_ran & 0 & -0.893, L & vr\_ga2 & vo\_ran & 0.013 & 0.614, L \\ 
        vr\_ga2 & vo\_ga1 & 0 & -0.886, L & vr\_ga2 & vo\_ga1 & 0.001 & 0.814, L \\ 
        vr\_ga2 & vo\_ga2 & 0 & -1.0, L & vr\_ga2 & vo\_ga2 & 0.057 & 0.429, M \\ 
        vo\_ran & vo\_ga1 & 0.052 & -0.52, L & vo\_ran & vo\_ga1 & 0.045 & 0.54, L \\ 
        vo\_ran & vo\_ga2 & 0 & -1.0, L & vo\_ran & vo\_ga2 & 0.578 & -0.143, N \\ 
        vo\_ga1 & vo\_ga2 & 0.004 & -0.7, L & vo\_ga1 & vo\_ga2 & 0.015 & -0.6, L \\ \hline
    \end{tabular}}
    \label{tab:rq2-uav}
\end{table}
\begin{table}[!ht]
    \centering
    \caption{Number of revealed failures and their sparseness in UC2}
    \scalebox{0.9}{
    \begin{tabular}{cccc|cccc}
    \hline
        Number of failures & ~ & ~ & ~ & Failure sparseness & ~ & ~ & ~ \\ \hline
        A & B & p-value & Effect size & A & B & p-value & Effect size \\ 
        ran & ga1 & 0.343 & -0.26, S & ran & ga1 & 0.345 & 0.26, S \\ 
        ran & ga2 & 0.303 & 0.257, S & ran & ga2 & 0.578 & 0.143, N \\ 
        ran & vr\_ran & 0.011 & 0.65, L & ran & vr\_ran & 0.448 & 0.2, S \\ 
        ran & vr\_ga1 & 0.721 & 0.113, N & ran & vr\_ga1 & 0.515 & -0.2, S \\ 
        ran & vr\_ga2 & 0 & -0.98, L & ran & vr\_ga2 & 1 & 0.0, N \\ 
        ran & vo\_ran & 0 & -1.0, L & ran & vo\_ran & 0.212 & -0.34, M \\ 
        ran & vo\_ga1 & 0 & -1.0, L & ran & vo\_ga1 & 0.275 & -0.291, S \\ 
        ran & vo\_ga2 & 0 & -1.0, L & ran & vo\_ga2 & 0.473 & 0.2, S \\ 
        ga1 & ga2 & 0.048 & 0.436, M & ga1 & ga2 & 0.93 & -0.029, N \\ 
        ga1 & vr\_ran & 0.004 & 0.725, L & ga1 & vr\_ran & 0.767 & -0.083, N \\ 
        ga1 & vr\_ga1 & 0.476 & 0.212, S & ga1 & vr\_ga1 & 0.173 & -0.4, M \\ 
        ga1 & vr\_ga2 & 0.001 & -0.88, L & ga1 & vr\_ga2 & 0.521 & -0.18, S \\ 
        ga1 & vo\_ran & 0 & -1.0, L & ga1 & vo\_ran & 0.054 & -0.52, L \\ 
        ga1 & vo\_ga1 & 0 & -1.0, L & ga1 & vo\_ga1 & 0.084 & -0.455, M \\ 
        ga1 & vo\_ga2 & 0 & -1.0, L & ga1 & vo\_ga2 & 0.623 & -0.14, N \\ 
        ga2 & vr\_ran & 0.044 & 0.47, M & ga2 & vr\_ran & 0.939 & -0.024, N \\ 
        ga2 & vr\_ga1 & 0.656 & -0.125, N & ga2 & vr\_ga1 & 0.33 & -0.268, S \\ 
        ga2 & vr\_ga2 & 0 & -0.993, L & ga2 & vr\_ga2 & 0.838 & -0.057, N \\ 
        ga2 & vo\_ran & 0 & -1.0, L & ga2 & vo\_ran & 0.084 & -0.429, M \\ 
        ga2 & vo\_ga1 & 0 & -1.0, L & ga2 & vo\_ga1 & 0.198 & -0.312, S \\ 
        ga2 & vo\_ga2 & 0 & -1.0, L & ga2 & vo\_ga2 & 0.93 & -0.029, N \\ 
        vr\_ran & vr\_ga1 & 0.03 & -0.594, L & vr\_ran & vr\_ga1 & 0.157 & -0.396, M \\ 
        vr\_ran & vr\_ga2 & 0 & -1.0, L & vr\_ran & vr\_ga2 & 0.767 & -0.083, N \\ 
        vr\_ran & vo\_ran & 0 & -1.0, L & vr\_ran & vo\_ran & 0.07 & -0.467, M \\ 
        vr\_ran & vo\_ga1 & 0 & -1.0, L & vr\_ran & vo\_ga1 & 0.117 & -0.394, M \\ 
        vr\_ran & vo\_ga2 & 0 & -1.0, L & vr\_ran & vo\_ga2 & 0.974 & -0.017, N \\ 
        vr\_ga1 & vr\_ga2 & 0.001 & -0.95, L & vr\_ga1 & vr\_ga2 & 0.36 & 0.275, S \\ 
        vr\_ga1 & vo\_ran & 0 & -1.0, L & vr\_ga1 & vo\_ran & 0.573 & -0.175, S \\ 
        vr\_ga1 & vo\_ga1 & 0 & -1.0, L & vr\_ga1 & vo\_ga1 & 0.6 & -0.159, S \\ 
        vr\_ga1 & vo\_ga2 & 0 & -1.0, L & vr\_ga1 & vo\_ga2 & 0.173 & 0.4, M \\ 
        vr\_ga2 & vo\_ran & 0 & -1.0, L & vr\_ga2 & vo\_ran & 0.14 & -0.4, M \\ 
        vr\_ga2 & vo\_ga1 & 0 & -1.0, L & vr\_ga2 & vo\_ga1 & 0.275 & -0.291, S \\ 
        vr\_ga2 & vo\_ga2 & 0 & -1.0, L & vr\_ga2 & vo\_ga2 & 0.734 & 0.1, N \\ 
        vo\_ran & vo\_ga1 & 0.377 & -0.236, S & vo\_ran & vo\_ga1 & 0.751 & 0.091, N \\ 
        vo\_ran & vo\_ga2 & 0.005 & -0.76, L & vo\_ran & vo\_ga2 & 0.021 & 0.62, L \\ 
        vo\_ga1 & vo\_ga2 & 0.034 & -0.555, L & vo\_ga1 & vo\_ga2 & 0.073 & 0.473, M \\ \hline
    \end{tabular}}
    \label{tab:rq2-ads}
\end{table}

\end{document}